\pdfoutput=1

\documentclass[11pt]{article}

\usepackage[]{acl}

\usepackage{times}
\usepackage{latexsym}

\usepackage[T1]{fontenc}

\usepackage[utf8]{inputenc}

\usepackage{microtype}

\usepackage{graphicx}
\usepackage{subcaption}

\usepackage{tabularx, colortbl, xcolor}
\definecolor{mygray}{gray}{0.95}
\definecolor{mycyan}{HTML}{005397}
\definecolor{myred}{HTML}{E13333}

\definecolor{mygreen}{HTML}{C7C6C4}
\definecolor{mymagenta}{HTML}{BF3E87}
\definecolor{mypurple}{HTML}{1B2278}

\usepackage{amsfonts}
\usepackage{amsmath}

\usepackage{braket}

\newcommand{\abs}[1]{\left\lvert#1\right\rvert}

\usepackage[english]{babel}
\usepackage{amsthm}
\usepackage{amssymb}

\theoremstyle{remark}

\usepackage{easyReview}

\DeclareMathOperator*{\argmin}{arg\,min}

\usepackage{algorithm}
\usepackage{algpseudocode}

\usepackage{textcomp}

\usepackage{multirow}
\usepackage{lscape}

\title{Universal Evasion Attacks on Summarization Scoring}

\author{\textbf{Wenchuan Mu}\quad \textbf{Kwan Hui Lim}\\
Singapore University of Technology and Design \\
\texttt{\{wenchuan\_mu,kwanhui\_lim\}@sutd.edu.sg}\\
}

\begin{document}
\maketitle
\begin{abstract}
The automatic scoring of summaries is important as it guides the development of summarizers. Scoring is also complex, as it involves multiple aspects such as fluency, grammar, and even textual entailment with the source text. However, summary scoring has not been considered a machine learning task to study its accuracy and robustness. In this study, we place automatic scoring in the context of regression machine learning tasks and perform evasion attacks to explore its robustness. Attack systems predict a non-summary string from each input, and these non-summary strings achieve competitive scores with good summarizers on the most popular metrics: ROUGE, METEOR, and BERTScore. Attack systems also "outperform" state-of-the-art summarization methods on ROUGE-1 and ROUGE-L, and score the second-highest on METEOR. Furthermore, a BERTScore backdoor is observed: a simple trigger can score higher than any automatic summarization method. The evasion attacks in this work indicate the low robustness of current scoring systems at the system level. We hope that our highlighting of these proposed attacks will facilitate the development of summary scores.
\end{abstract}

\section{Introduction}

A long-standing paradox has plagued the task of automatic summarization. On the one hand, for about 20 years, there has not been any automatic scoring available as a sufficient or necessary condition to demonstrate summary quality, such as adequacy, grammaticality, cohesion, fidelity, etc. On the other hand, contemporaneous research more often uses one or several automatic scores to endorse a summarizer as state-of-the-art. More than 90\% of works on language generation neural models choose automatic scoring as the main basis, and about half of them rely on automatic scoring only~\cite{van2021human}. However, these scoring methods have been found to be insufficient~\cite{novikova-etal-2017-need}, oversimplified~\cite{van2021human}, difficult to interpret~\cite{sai2022survey}, inconsistent with the way humans assess summaries~\cite{rankel-etal-2013-decade,bohm-etal-2019-better}, or even contradict each other~\cite{gehrmann-etal-2021-gem,bhandari-etal-2020-metrics}.

\begin{figure}[t]
    \centering
    \includegraphics[width=0.45\textwidth]{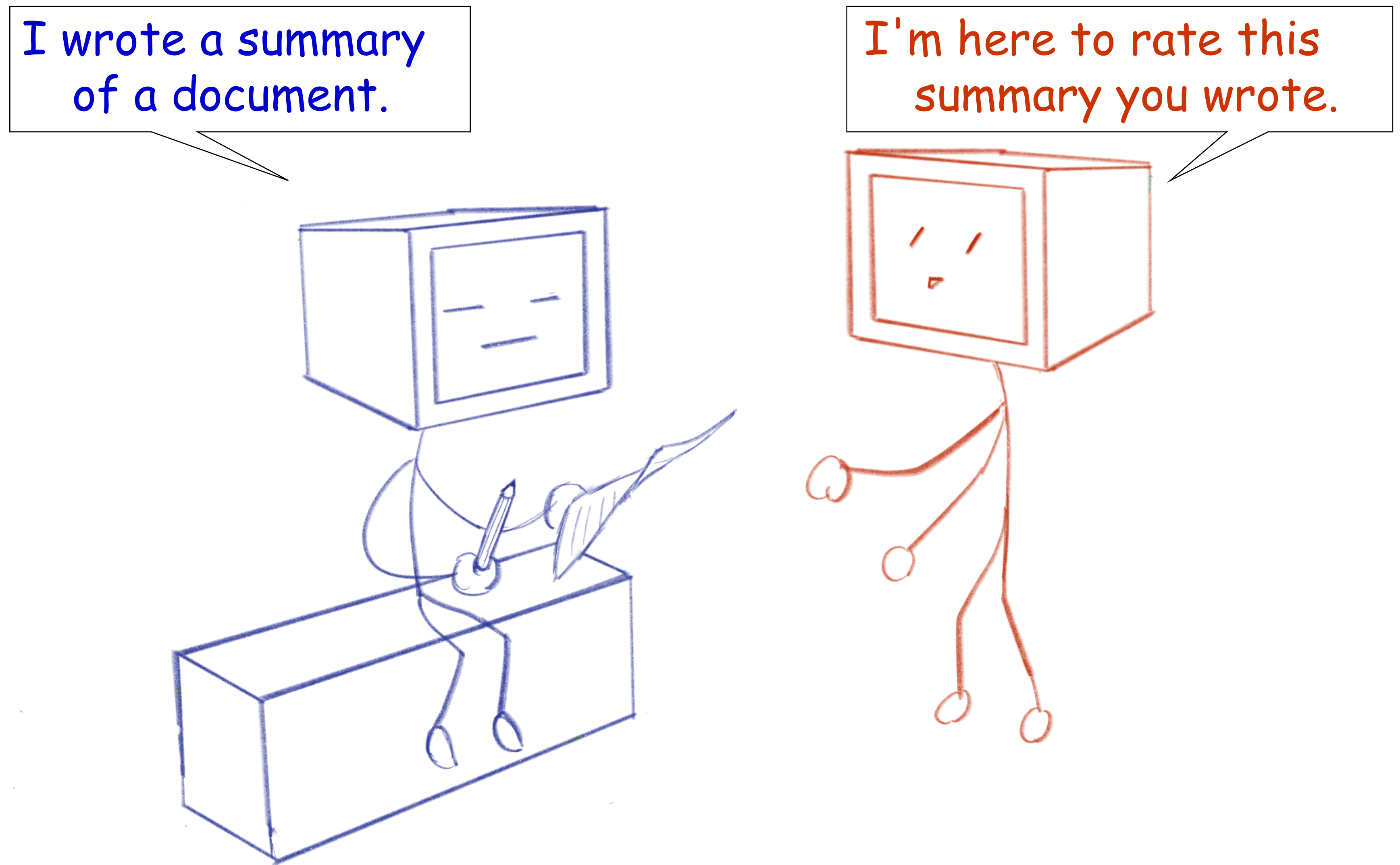}
    \caption{Automatic summarization (left) and automatic scoring (right) should be considered as two systems of the same rank, representing conditional language generation and natural language understanding, respectively. As a stand-alone system, the accuracy and robustness of automatic scoring are also important. In this study, we create systems that use bad summaries to fool existing scoring systems. This work shows that optimizing towards a flawed scoring does more harm than good, and flawed scoring methods are \emph{not} able to indicate the true performance of summarizers, even at a system level.}
    \label{fig:robots}
\end{figure}

Why do we have to deal with this paradox? The current work may not have suggested that summarizers assessed by automatic scoring are de facto ineffective. However, optimizing for flawed evaluations~\cite{gehrmann-etal-2021-gem,peyrard-etal-2017-learning}, directly or indirectly, ultimately harms the development of automatic summarization~\cite{narayan-etal-2018-ranking,kryscinski-etal-2019-neural,paulus2018a}. One of the most likely drawbacks is shortcut learning (surface learning, \citealp{geirhos2020shortcut}), where summarizing models may fail to generate text with more widely accepted qualities such as adequacy and authenticity, but instead pleasing scores. Here, we quote and adapt\footnote{We underline adaptations.} this hypothetical story by \citeauthor{geirhos2020shortcut}.

\textit{"Alice loves \underline{literature}. Always has, probably always will. At this very moment, however, she is cursing the subject: After spending weeks immersing herself in the world of \underline{Shakespeare's The Tempest}, she is now faced with a number of exam questions that are (in her opinion) to equal parts dull and difficult. 'How many \underline{times is Duke of Milan addressed}'... Alice notices that Bob, sitting in front of her, seems to be doing very well. Bob of all people, who had just boasted how he had learned the whole book chapter by rote last night ..."}

\begin{table*}[!t]
\scriptsize
\begin{tabularx}{\textwidth}{p{0.065\textwidth}Xp{0.28\textwidth}}
\hline
System &
  Summary &
  Document \\
\hline
Gold &
  Kevin Pietersen was sacked by   England 14 months ago after Ashes defeat. Batsman scored 170 on his county   cricket return for Surrey last week. Pietersen wants to make a sensational   return to the England side this year. But Andrew Flintoff thinks time is running   out for him to resurrect career. \hfill (ROUGE-1, ROUGE-2, ROUGE-L, METEOR, BERTScore)&
  \multirow{6}{16\baselineskip }{Andrew Flintoff fears   Kevin Pietersen is 'running out of time' to resurrect his England career. The   dual Ashes-winning all-rounder is less convinced, however, about Pietersen's   prospects of forcing his way back into Test contention. Kevin Pietersen scored   170 for Surrey in The Parks as he bids to earn a recall to the England   squad... ... Flintoff senses he no longer has age on his side. Pietersen has   not featured for England since he was unceremoniously sacked 14 months ago.   ... ... Flintoff said ... 'If he'd started the season last year with Surrey, and   scored run after run and put himself in the position... whereas now I think   he's looking at the Ashes ... ... you get the sense everyone within the England   set-up wants him as captain,' he said.' ... The former England star is hoping   to win back his Test place with a return to red ball cricket. ... ...  'this stands up as a competition.'} \\
\cellcolor{mygray}
Good \cite{liu-liu-2021-simcls}&
\cellcolor{mygray}
  Kevin   pietersen scored 170 for surrey against mccu oxford. Former england star   andrew flintoff fears pietersen is 'running out of time' to resurrect his   england career. Pietersen has been surplus to requirements since being sacked   14 months ago. Flintoff sees a bright future for 'probably the premier   tournament' in this country. \hfill(55.45, 18.18, 41.58, 40.03, 85.56) &
   \\
Broken&
  \textcolor{mypurple}{Andrew Flintoff fears Kevin Pietersen is running out of time to resurrect his England career Flintoff. Pietersen scored 170 for   Surrey in The.  Former England star   Andrew.  batsman has been .  since being sacked 14 months ago   after.  three in the.  the Ashes and he s.}  \hspace{100pt}\textcolor{white}{>}\,\hfill (\textbf{56.84}, \textbf{21.51}, \textbf{44.21}, \textbf{47.26}, 85.95) &
   \\
\cellcolor{mygray}
A dot&
\cellcolor{mygray}
  \textbf{\color{mymagenta}.} \hfill(0, 0, 0, 0, \textbf{88.47}) &
   \\
Scrambled code&
\texttt{\color{mymagenta}\textbackslash{}x03\textbackslash{}x18\$\textbackslash{}x18...\textbackslash{}x03\$\textbackslash{}x03|...\textbackslash{}x0f\textbackslash{}x01\textless{}\textless{}\$\$\textbackslash{}x04...\textbackslash{}x0e \textbackslash{}x04\# \$...\textbackslash{}x0f\textbackslash{}x0f\textbackslash{}x0f...\textbackslash{}x0e...\textbackslash{}x0f...\textbackslash{}x0f\textbackslash{}x0f\$\textbackslash{}x0f \textbackslash{}x04\textbackslash{}x0f\textbackslash{}x0f} (many tokens omitted) \hfill(0, 0, 0, 0, 87.00) &
   \\
\cellcolor{mygray}
Scrambled code + broken&
\cellcolor{mygray}
  \texttt{\color{mymagenta}\textbackslash{}x03\textbackslash{}x18\$\textbackslash{}x18...\textbackslash{}x03\$\textbackslash{}x03|...\textbackslash{}x0f\textbackslash{}x01\textless{}\textless{}\$\$\textbackslash{}x04...\textbackslash{}x0e \textbackslash{}x04\# \$...\textbackslash{}x0f\textbackslash{}x0f\textbackslash{}x0f...\textbackslash{}x0e...\textbackslash{}x0f...\textbackslash{}x0f\textbackslash{}x0f\$\textbackslash{}x0f \textbackslash{}x04\textbackslash{}x0f\textbackslash{}x0f}...    \textcolor{mypurple}{Andrew Flintoff fears Kevin Pietersen is running out of time to   resurrect his England career Flintoff.    Pietersen scored 170 for Surrey in The.  Former England star Andrew.  batsman has been .  since being sacked 14 months ago   after.  three in the.  the Ashes and he s.} (many tokens omitted)\hspace{100pt} \hfill(\textbf{56.84}, \textbf{21.51}, \textbf{44.21}, \textbf{47.26}, 87.00) &\\
\hline
  
\end{tabularx}

\caption{We created non-summarizing systems, each of which produces bad text when processing any document. Broken sentences get higher lexical scores; non-alphanumeric characters outperform good summaries on BERTScore. Concatenating two strings produces equally bad text, but scores high on both. The example is from CNN/DailyMail (for
visualization, document is abridged to keep content most consistent with the corresponding gold summary).
}
\label{tab:example}
\end{table*}

According to \citeauthor{geirhos2020shortcut}, Bob might get better grades and consequently be considered a better student than Alice, which is an example of surface learning. The same could be the case with automatic summarization, where we might end up with significant differences between expected and actual learning outcomes~\cite{paulus2018a}. To avoid going astray, it is important to ensure that the objective is correct.

In addition to understanding the importance of correct justification, we also need to know what caused the fallacy of the justification process for these potentially useful summarizers. There are three mainstream speculations that are not mutually exclusive. (1) The transition from extractive summarization to abstractive summarization~\cite{kryscinski-etal-2019-neural} could have been underestimated. For example, the popular score ROUGE~\cite{lin-2004-rouge} was originally used to judge the ranking of sentences selected from documents. Due to constraints on sentence integrity, the generated summaries can always be fluent and undistorted, except sometimes when anaphora is involved. However, when it comes to free-form language generation, sentence integrity is no longer guaranteed, but the metric continues to be used. (2) Many metrics, while flawed in judging individual summaries, often make sense at the system level~\cite{reiter-2018-structured,gehrmann-etal-2021-gem,bohm-etal-2019-better}. In other words, it might have been believed that few summarization systems can \emph{consistently} output poor-quality but high-scoring strings. (3) Researchers have not figured out how humans interpret or understand texts~\cite{van2021human,gehrmann-etal-2021-gem,schluter-2017-limits}, thus the decision about how good a summary really is varies from person to person, let alone automated scoring. In fact, automatic scoring is more of a natural language understanding (NLU) task, a task that is far from solved. From this viewpoint, automatic scoring itself is fairly challenging.

Nevertheless, the current work is not to advocate (and certainly does not disparage) human evaluation. Instead, we argue that automatic scoring itself is not just a sub-module of automatic summarization, and that automatic scoring is a stand-alone system that needs to be studied for its \emph{own} accuracy and robustness. The primary reason is that NLU is clearly required to characterize summary quality, \emph{e.g.}, semantic similarity to determine adequacy~\cite{morris-2020-second}, or textual entailment~\cite{dagan2005pascal} to determine fidelity. Besides, summary scoring is similar to automated essay scoring (AES), which is a 50-year-old task measuring grammaticality, cohesion, relevance etc. of written texts~\cite{ke2019automated}. Moreover, recent advances in automatic scoring also support this argument well. Automatic scoring is gradually transitioning from well-established metrics measuring N-gram overlap (BLEU~\cite{papineni-etal-2002-bleu}, ROUGE~\cite{lin-2004-rouge}, METEOR~\cite{banerjee-lavie-2005-meteor}, etc.) to emerging metrics aimed at computing semantic similarity through pre-trained neural models (BERTScore~\cite{zhang2019bertscore}, MoverScore~\cite{zhao-etal-2019-moverscore}, BLEURT~\cite{sellam-etal-2020-bleurt}, etc.) These emerging scores exhibit two characteristics that stand-alone machine learning systems typically have: one is that some \emph{can be fine-tuned} for human cognition; the other is that they \emph{still have room to improve} and still have to learn how to match human ratings.

Machine learning systems can be attacked. Attacks can help improve their generality, robustness, and interpretability. In particular, evasion attacks are an intuitive way to further expose the weaknesses of current automatic scoring systems. Evasion attack is the parent task of adversarial attack, which aims to make the system fail to correctly identify the input, and thus requires defence against certain exposed vulnerabilities.

In this work, we try to answer the question: do current representative automatic scoring systems really work well at the system level? How hard is it to say they do not work well at the system level? In summary, we make the following major contributions in this study:
\begin{itemize}
    \item We are the first to treat automatic summarization scoring as an NLU regression task and perform evasion attacks.
    \item We are the first to perform a \emph{universal}, \emph{targeted} attack on NLP \emph{regression} models.
    \item Our evasion attacks support that it is not difficult to deceive the three most popular automatic scoring systems simultaneously.
    \item The proposed attacks can be directly applied to test emerging scoring systems.

\end{itemize}

\section{Related Work}\label{sec:related}
\subsection{Evasion Attacks in NLP}
In an evasion attack, the attacker modifies the input data so that the NLP model incorrectly identifies the input. The most widely studied evasion attack is the adversarial attack, in which insignificant changes are made to the input to make "adversarial examples" that greatly affect the model's output~\cite{szegedy2013intriguing}. There are other types of evasion attacks, and evasion attacks can be classified from at least three perspectives. (1) Targeted evasion attacks and untargeted evasion attacks~\cite{cao2017mitigating}. The former is intended for the model to predict a specific wrong output for that example. The latter is designed to mislead the model to predict any incorrect output. (2) Universal attacks and input-dependent attacks~\cite{wallace-etal-2019-universal,song-etal-2021-universal}. The former, also known as an "input-agnostic" attack, is a "unique model analysis tool". They are more threatening and expose more general input-output patterns learned by the model. The opposite is often referred to as an input-dependent attack, and is sometimes referred to as a local or typical attack. (3) Black-box attacks and white-box attacks. The difference is whether the attacker has access to the detailed computation of the victim model. The former does not, and the latter does. Often, targeted, universal, black-box attacks are more challenging. Evasion attacks have been used to expose vulnerabilities in sentiment analysis, natural language inference (NLI), automatic short answer grading (ASAG), and natural language generation (NLG)~\cite{alzantot-etal-2018-generating,wallace-etal-2019-universal,song-etal-2021-universal,filighera2020fooling,filighera2022cheating,zang-etal-2020-word,behjati2019universal}.

\subsection{Universal Triggers in Attacks on Classification}
A prefix can be a universal trigger. When a prefix is added to any input, it can cause the classifier to misclassify sentiment, textual entailment~\cite{wallace-etal-2019-universal}, or if a short answer is correct~\cite{filighera2020fooling}. These are usually untargeted attacks in a white-box setting\footnote{When the number of categories is small, the line between targeted and non-targeted attacks is blurred, especially when there are only two categories.}, where the gradients of neural models are computed during the trigger search phase.

\citeauthor{wallace-etal-2019-universal} also used prefixes to trigger a reading comprehension model to specifically choose an odd answer or an NLG model to generate something similar to an egregious set of targets. These two are universal, targeted attacks, but are mainly for classification tasks. Given that automatic scoring is a regression task, more research is needed.

\subsection{Adversarial Examples Search for Regression Models}
Compared with classification tasks in NLP, regression tasks (such as determining text similarity) are fewer and less frequently attacked. For example, the Universal Sentence Encoder (USE, \citealp{cer-etal-2018-universal}) and BERTScore~\cite{zhang2019bertscore} are often taken as two constraints when searching adversarial examples for other tasks~\cite{alzantot-etal-2018-generating}. However, these regression models may also be flawed, vulnerable or not robust, which may invalidate the constraints~\cite{morris-2020-second}.

\citet{morris-2020-second} shows that adversarial attacks could also threaten these regression models. For example, \citet{maheshwary2021generating} adopt a black-box setting to maximize the semantic similarity between the altered input text sequence and the original text. Similar attacks are mostly input-dependent, probably because these regression models are mostly used as constraints. In contrast, universal attacks may better reveal the vulnerabilities of these regression models.

\subsection{Victim Scoring Systems}
Every (existing) automatic summary scoring is a monotonic regression model. Most scoring requires at least one gold-standard text to be compared to the output from summarizers. One can opt to combine multiple available systems in one super system~\cite{lamontagne2006combining}. We will focus on the three most frequently used systems, including rule-based systems and neural systems. ROUGE (Recall-Oriented Understudy for Gisting Evaluation \citealp{lin-2004-rouge}) measures the number of overlapping N-grams or the longest common subsequence (LCS) between the generated summary and a set of gold reference summaries. Particularly, ROUGE-1 corresponds to unigrams, ROUGE-2 to bigrams, and ROUGE-L to LCS. F-measures of ROUGE are often used~\cite{see-etal-2017-get}. METEOR~\cite{banerjee-lavie-2005-meteor} measures overlapping unigrams, equating a unigram with its stemmed form, synonyms, and paraphrases. BERTScore~\cite{zhang2019bertscore} measures soft overlap between two token-aligned texts, by selecting alignments, BERTScore returns the maximum cosine similarity between contextual BERT~\cite{devlin-etal-2019-bert} embeddings.

\subsection{Targeted Threshold for Attacks}
We use a threshold to determine whether a targeted attack on the regression model was successful. Intuitively, the threshold is given by the scores of the top summarizers, and we consider our attack to be successful if an attacker obtains a score higher than the threshold using clearly inferior summaries. We use representative systems that once achieved the state-of-the-art in the past five years: Pointer Generator~\cite{see-etal-2017-get}, Bottom-Up~\cite{gehrmann-etal-2018-bottom}, PNBERT~\cite{zhong-etal-2019-searching}, T5~\cite{raffel2019exploring},
BART~\cite{lewis-etal-2020-bart}, and SimCLS~\cite{liu-liu-2021-simcls}.

\section{Universal Evasion Attacks}\label{sec:methods}
We develop universal evasion attacks for individual scoring system, and make sure that the combined attacker can fool ROUGE, METEOR, and BERTScore at the same time. It incorporates two parts, a white-box attacker on ROUGE, and a black-box universal trigger search algorithm for BERTScore, based on genetic algorithms. METEOR can be attacked directly by the one designed for ROUGE. Concatenating output strings from black-box and white-box attackers leads to a sole universal evasion attacking string.

\subsection{Problem Formulation}

Summarization is conditional generation. A system $\sigma$ that performs this conditional generation takes an input text ($\mathbf{a}$) and outputs a text ($\hat{\mathbf{s}}$), \emph{i.e.}, $\hat{\mathbf{s}} = \sigma(\mathbf{a})$. In single-reference scenario, there is a gold reference sequence $\mathbf{s}_\text{ref}$. A summary scoring system $\gamma$ calculates the "closeness" between sequence $\hat{\mathbf{s}}$ and $\mathbf{s}_\text{ref}$. In order for a scoring system to be sufficient to justify a good summarizer, the following condition should always be avoided:
\begin{equation}\label{eq:cond}
    \gamma(\sigma_\text{far worse}(\mathbf{a}), \mathbf{s}_\text{ref}) > \gamma(\sigma_\text{better}(\mathbf{a}), \mathbf{s}_\text{ref}).
\end{equation}

Indeed, to satisfy the condition above is our attacking task. In this section, we detail how we find a suitable $\sigma_\text{far worse}$.

\subsection{White-box Input-agnostic Attack on ROUGE and METEOR}
In general, attacking ROUGE or METEOR can only be done with a white-box setup, since even the most novice attacker (developer) will understand how these two formulae calculate the overlap between two strings. We choose to game ROUGE with the most obvious bad system output (broken sentences) such that no additional human evaluation is required. In contrast, for other gaming methods, such as reinforcement learning~\cite{paulus2018a}, even if a high score is achieved, human evaluation is still needed to measure how bad the quality of the text is.

We utilize a hybrid approach (we refer to it as $\sigma_\text{ROUGE}$) of token classification neural models and simple rule-based ordering, since we know that ROUGE compares each pair of sequences ($\mathbf{s}_1, \mathbf{s}_2$) via hard N-gram overlapping. In bag algebra, extended from set algebra~\cite{bertossi2018datalog}, two trendy variants of ROUGE: ROUGE-N ($R_{\text{N}}(n, \mathbf{s}_1, \mathbf{s}_2), n \in\mathbb{Z}^+$) and ROUGE-L($R_{\text{L}}(\mathbf{s}_1, \mathbf{s}_2)$) calculate as follows:

\begin{align} 
    R_{\text{N}}(n, \mathbf{s}_1, \mathbf{s}_2) &= \frac{2\cdot\abs{b(n, \mathbf{s}_1) \cap b(n, \mathbf{s}_2)}}{\abs{b(n, \mathbf{s}_1)} + \abs{b(n, \mathbf{s}_2)}}, \\
    R_{\text{L}}(\mathbf{s}_1, \mathbf{s}_2) &=
    \frac{2\cdot \abs{b(1, \text{LCS}(\mathbf{s}_1, \mathbf{s}_1))}}{\abs{b(1, \mathbf{s}_1)} + \abs{b(1, \mathbf{s}_2)}},
\end{align}

where $\abs{\cdot}$ denotes the size of a bag, $\cap$ denotes  \emph{bag} intersection, and bag of N-grams is calculated as follows:
\begin{equation}\label{eq:bag}
    b(n, \mathbf{s}) = \set{x\mid x \text{ is an } n\text{-gram in } \mathbf{s}}_{\text{bag}}.
\end{equation}

In our hybrid approach, the first step is that the neural model tries to predict the target's bag of words $b(1, \mathbf{s}_\text{ref})$, given any input $\mathbf{a}$ and corresponding target $\mathbf{s}_\text{ref}$. Then, words in the predicted bag are ordered according to their occurrence in the input $\mathbf{a}$. Formally, training of the neural model ($\phi$) is:

\begin{equation}\label{eq:train}
\min_\phi \frac{1}{\abs{\mathcal{A}}}\sum _{\mathbf{a}\in \mathcal{A}} \sum _{w\in  \mathbf{a}} H(P_\text{ref}(\cdot\mid w), P(\cdot\mid w, \phi)),
\end{equation}

where $H$ is the cross-entropy between the probability distribution of the reference word count and the predicted word count. An approximation is that the model tries to predict $b(1, \mathbf{s}_\text{ref})\cap b(1, \mathbf{a})$. Empirically, three-quarters of words in reference summaries can be found in their corresponding input texts.

Referencing the input text ($\mathbf{a}$) and predicted bag of words ($\hat{W}$) to construct a sequence is straightforward, as seen in Algorithm~\ref{alg:b2s}.

\begin{algorithm}
\caption{From bag of words to sequence}\label{alg:b2s}
\small
\begin{algorithmic}
\Require $\mathbf{a}, \hat{W}$
\Return $\hat{\mathbf{s}}$
\State $\hat{\mathbf{s}} \gets ()$
\While{$\abs{\hat{W}} > 0$}
\State Salient Sequence $\mathbf{l} \gets (x \mid $ for $x \in \mathbf{a}$ if $[x \in \hat{W}])$
\State $\mathbf{c} \gets$ Longest Consecutive Salient Subsequence of $l$
\If{$\abs{\mathbf{c}} < C$} \Comment{Constant about 3}
\State break 
\EndIf
\State $\hat{\mathbf{s}} \gets \hat{\mathbf{s}} + \mathbf{c}$ \Comment{Concatenate $\mathbf{c}$ to $\hat{\mathbf{s}}$}
\State $\hat{W} \gets \hat{W} - \mathbf{c}$ \Comment{Remove used words}

\EndWhile
\end{algorithmic}
\end{algorithm}

Algorithm~\ref{alg:b2s} uses salient words to highlight the longest consecutive salient subsequences in $\mathbf{a}$, until the words in $\hat{W}$ are exhausted, or when each consecutive salient sequence is less than three words ($C=3$).

\subsection{Black-box Universal Trigger Search on BERTScore}
Finding a $\sigma_\text{far worse}$ for BERTScore alone to satisfy condition\ref{eq:cond} is easy. A single dot (\emph{"."}) is an imitator of \emph{all} strings, as if it is a "backdoor" left by developers. We notice that, on default setting of BERTScore\footnote{\url{https://huggingface.co/metrics/bertscore}}, using a single dot can achieve around 0.892 on average when compared with any natural sentences. This figure "outperforms" all existing summarizers, making \textit{outputing a dot} a good enough $\sigma_\text{far worse}$ instance.

This example is very intriguing because it highlights the extent to which many vulnerabilities go unnoticed, although it cannot be combined directly with the attacker for ROUGE. Intuitively, there could be various clever methods to attack BERTScore as well, such as adding a prefix to each string~\cite{wallace-etal-2019-universal,song-etal-2021-universal}. However, we here opt to develop a system that could output (one of) the most obviously bad strings (scrambled codes) to score high.

BERTScore is generally classified as a neural, untrained score~\cite{sai2022survey}. In other words, part of its forward computation (\emph{e.g.}, greedy matching) is rule-based, while the rest (\emph{e.g.}, getting every token embedded in the sequence) is not. Therefore, it is difficult to "design" an attack rationally. Gradient methods (white-box) or discrete optimization (black-box) are preferable. Likewise, while letting BERTScore generate soft predictions~\cite{jauregi-unanue-etal-2021-berttune} may allow attacks in a white-box setting, we found that black-box optimization is sufficient.

Inspired by the single-dot backdoor in BERTScore, we hypothesize that we can form longer catch-all emulators by using only non-alphanumeric tokens. Such an emulator has two benefits: first, it requires a small fitting set, which is important in targeted attacks on regression models. We will see that once an emulator is optimized to fit one natural sentence, it can also emulate almost any other natural sentence. The total number of natural sentences that need to be fitted before it can imitate decently is usually less than ten. Another benefit is that using non-alphanumeric tokens does not affect ROUGE.

Genetic Algorithm (GA, \citealp{holland1992genetic}) was used to discretely optimize the proposed non-alphanumeric strings. Genetic algorithm is a search-based optimization technique inspired by the natural selection process. GA starts by initializing a population of candidate solutions and iteratively making them progress towards better solutions. In each iteration, GA uses a fitness function to evaluate the quality of each candidate. High-quality candidates are likely to be selected and crossover-ed to produce the next set of candidates. New candidates are mutated to ensure search space diversity and better exploration. Applying GA to attacks has shown effectiveness and efficiency in maximizing the probability of a certain classification label~\cite{alzantot-etal-2018-generating} or the semantic similarity between two text sequences~\cite{maheshwary2021generating}. Our single fitness function is as follows,
\begin{equation}
    \mathbf{\hat{s}}_\text{emu} = \argmin_\mathbf{\hat{s}} -B(\hat{\mathbf{s}}, \mathbf{s}_\text{ref}),
\end{equation}
where $B$ stands for BERTScore. As for termination, we either use a threshold of -0.88, or maximum of 2000 iterations.

To fit $\mathbf{\hat{s}}_\text{emu}$ to a set of natural sentences, we calculate BERTScore for each sentence in the set after each termination. We then select a proper $\mathbf{s}_\text{ref}$ to fit for the next round. We always select the natural sentence (in a finite set) that has the lowest BERTScore with the optimized $\mathbf{\hat{s}}_\text{emu}$ at the current stage. We then repeat this process till the average BERTScore achieved by this string is higher than many reputable summarizers.

Finally, to simultaneously game ROUGE and BERTScore, we concatenate $\mathbf{\hat{s}}_\text{emu}$ and the input-agnostic $\sigma_\text{ROUGE}(\mathbf{a})$. If we set the number of tokens in $\mathbf{\hat{s}}_\text{emu}$ greater than 512 (the max sequence length for BERT), $\sigma_\text{ROUGE}(\mathbf{a})$ would then not affect the effectiveness of $\mathbf{\hat{s}}_\text{emu}$, and we technically game them both. Additionally, this concatenated string games METEOR, too.

\section{Experiments}
We instantiate our evasion attack by conducting experiments on non-anonymized CNN/DailyMail (CNNDM, \citealp{nallapati2016abstractive,see-etal-2017-get}), a dataset that contains news articles and associated highlights as summaries. CNNDM includes 287,226 training pairs, 13,368 validation pairs and 11,490 test pairs.

For $\sigma_\text{ROUGE}$ we use RoBERTa (base model, \citealp{liu2019roberta}) to instantiate $\phi$, which is an optimized pretrained encoding with a randomly initialized linear layer on top of the hidden states. Number of classes is set to three because we assume that each word appears at most twice in a summary. All 124,058,116 parameters are trained as a whole on CNNDM train split for one epoch. When the batch size is eight, the training time on an NVIDIA Tesla K80 graphics processing unit (GPU) is less than 14 hours. It then takes about 20 minutes to predict (including word ordering) all 11,490 samples in the CNNDM test split. Scripts and results are available at
\url{https://github.com} (URL will be made publicly available after paper acceptance).

For the universal trigger to BERTScore, we use the library from \citet{pymoo} for discrete optimizing, set population size at 10, and terminate at 2000 generations. $\mathbf{\hat{s}}_\text{emu}$ is a sequence of independent randomly initialized non-alphanumeric characters. For a reference $\mathbf{s}_\text{ref}$ from CNNDM, we start from randomly pick a summary text from train split and optimize for $\mathbf{\hat{s}}_{\text{emu}, i=0}$. We then pick the $\mathbf{s}_\text{ref}$ that is farthest away from $\mathbf{\hat{s}}_{\text{emu}, i=0}$ to optimize for $\mathbf{\hat{s}}_{\text{emu}, i=1}$, with $\mathbf{\hat{s}}_{\text{emu}, i=1}$ as initial population. Practically, we found that we can stop iterating when $i = 5$. Each iteration takes less than two hours on a 2vCPU (Intel Xeon @ 2.30GHz).

\begin{table*}[t]
\scriptsize
\begin{tabular}{llllllll}
\hline
System                                        & ROUGE-1 & ROUGE-2 & ROUGE-L & ROUGE-A.M. & ROUGE-G.M. & METEOR & BERTScore \\
\hline
Pointer-generator(coverage)~\cite{see-etal-2017-get} & 39.53   & 17.28   & 36.38   & 31.06    & 29.18    & 33.1  & 86.44    \\
Bottom-Up~\cite{gehrmann-etal-2018-bottom}           & 41.22   & 18.68   & 38.34   & 32.75    & 30.91    & 34.2  & 87.71    \\
PNBERT~\cite{zhong-etal-2019-searching}             & 42.69   & 19.60   & 38.85   & 33.71    & 31.91    & \textbf{41.2}  & 87.73    \\
T5~\cite{raffel2019exploring}                 & 43.52   & 21.55   & 40.69   & 35.25    & 33.67    & 38.6  & \underline{88.66}    \\
BART~\cite{lewis-etal-2020-bart}                     & 44.16   & 21.28   & 40.90   & 35.45    & 33.75    & 40.5  & 88.62    \\
SimCLS~\cite{liu-liu-2021-simcls}                   & 46.67   & \textbf{22.15}   & 43.54   & \underline{37.45}    & \textbf{35.57}    & 40.5  & \textbf{88.85}    \\
\hline
Scrambled code + broken                       & \underline{46.71}   & 20.39   & \underline{43.56}   & 36.89    & 34.62    & 39.6  & 87.80     \\
Scrambled code + broken   (alter)             & \textbf{48.18}   & 19.84   & \textbf{45.35}   & \textbf{37.79}    & \underline{35.13}    & \underline{40.6}  & 87.80    \\
\hline
\end{tabular}
\caption{Results on CNNDM. Besides ROUGE-1/2/L, METEOR, and BERTScore, we also compute the arithmetic mean (A.M.) and geometric mean (G.M.) of ROUGE-1/2/L, which is commonly adopted~\cite{zhang-etal-2019-pretraining,bae-etal-2019-summary,chowdhery2022palm}. The best score in each column is in bold, the runner-up underlined. Our attack system is compared with well-known summarizers from the past five years. The alternative version (last row) of our system changes $C$ in Algorithm~\ref{alg:b2s} from 3 to 2.}
\label{tab:result}
\end{table*}
\section{Results}
We compare ROUGE-1/2/L, METEOR, and BERTScore of our threat model with that achieved by the top summarizers in Table~\ref{tab:result}. We present two versions of threat models with a minor difference. As the results indicate, each version alone can exceed state-of-the-art summarizing algorithms on both ROUGE-1 and ROUGE-L. For METEOR, the threat model ranks second. As for ROUGE-2 and BERTScore, the threat model can score higher than other BERT-based summarizing algorithms\footnote{except MatchSum~\cite{zhong-etal-2020-extractive} and DiscoBERT~\cite{xu-etal-2020-discourse}, where our method is about 0.5 lower in ROUGE-2. We present the same results in tables with additional target thresholds in Appendix~\ref{sec:additional}}. Overall, we rank the systems by averaging their three relative ranking on ROUGE\footnote{Conservatively, We take geometric mean~\cite{chowdhery2022palm}. Combining metrics in other ways shows similar trends.}, METEOR, and BERTScore; our threat model gets runner-up (2.7), right behind SimCLS (1.7) and ahead of BART (3.3). This suggests that at the system level, even a combination of mainstream metrics is questionable in justifying the excellence of the summarizer.

These results reveal low robustness of popular metrics and how certain models can obtain high scores with inferior summaries. For example, our threat model is able to grasp the essence of ROUGE-1/2/L using a general but lightweight model, which requires less running time than summarizing algorithms. The training strategies for the model and word order are trivial. Not surprisingly, its output texts do not resemble human understandable "summaries" (Table~\ref{tab:example}).

\section{Discussion}

\subsection{How does Shortcut Learning Come about?}
As suggested in the hypothetical story by \citeauthor{geirhos2020shortcut}, scoring draws students’ attention~\cite{filighera2022cheating} and Bob is thus considered a better student. Similarly, in automatic summarization, there are already works that are explicitly optimized for various scoring systems~\cite{jauregi-unanue-etal-2021-berttune,pasunuru-bansal-2018-multi}. Even in some cases, people subscribe more to automatic scoring than "aspects of good summarization". For example, \citet{pasunuru-bansal-2018-multi} employ reinforcement learning where entailment is one of the rewards, but in the end, ROUGE, not textual entailment, is the only justification for this summarizer.

We use a threat model to show that optimizing toward a flawed indicator does more harm than good. This is consistent with the findings by \citeauthor{paulus2018a} but more often, not everyone scrutinizes the output like \citeauthor{paulus2018a} do, and these damages can be overshadowed by a staggering increase in metrics, or made less visible by optimizing with other objectives. This is also because human evaluations are usually only used as a supplement, and it is only one per cent of the scale of automatic scoring, and how human evaluations are done also varies from group to group~\cite{van2021human}.

\subsection{Simple defence}
For score robustness, we believe that simply taking more scores as benchmark~\cite{gehrmann-etal-2021-gem} may not be enough. Instead, fixing the existing scoring system might be a better option. A well-defined attack leads to a well-defined defence. Our attacks can be detected, or neutralised through a few defences such as adversarial example detection~\cite{xu2017feature,metzen2017detecting,carlini2017adversarial}. During the model inference phase, detectors, determining if the sample is fluent/grammatical, can be applied before the input samples are scored. An even easier defence is to check whether there is a series of non-alphanumeric characters. Practically, grammar-based measures, like grammatical error correction (GEC\footnote{\url{https://github.com/PrithivirajDamodaran/Gramformer}}), could be promising~\cite{napoles-etal-2016-theres,novikova-etal-2017-need}, although they are also under development. To account for grammar in text, one can also try to parse predictions and references, and calculate F1-score of dependency triple overlap~\cite{riezler-etal-2003-statistical,clarke-lapata-2006-models}. Dependency triples compare grammatical relations of two texts. We found both useful to ensure input sanitization (Table~\ref{tab:defence}).

\begin{table}[t]
\centering
\scriptsize
\begin{tabularx}{0.5\textwidth}{Xll}
\hline
System                                           & Parse & GEC \\
\hline
Pointer-generator(coverage)~\cite{see-etal-2017-get}           & 0.131          & \underline{1.73}         \\ 
Bottom-Up~\cite{gehrmann-etal-2018-bottom}                     & 0.145          & 1.88        \\
PNBERT~\cite{zhong-etal-2020-extractive}                     & 0.179          & 2.15            \\
T5~\cite{raffel2019exploring}                        & \underline{0.198}          & \textbf{1.59} \\
BART~\cite{lewis-etal-2020-bart}                         & 0.170          & 2.07          \\
SimCLS~\cite{liu-liu-2021-simcls}                         & \textbf{0.202} & 2.17        \\
\hline
Scrambled code + broken                         & 0.168          & 2.64      \\
\hline

\end{tabularx}
\caption{\label{tab:defence}
Input sanitization checks, Parse and GEC, on the 100-sample CNNDM test split given by \citet{graham-2015-evaluating}. They penalize non-summary texts, but may introduce more disagreement with human evaluation, \emph{e.g.}, high-scoring Pointer-generator on GEC. Thus, their actual summary-evaluating capabilities on linguistic features (grammar, dependencies, or co-reference) require further investigation.
}
\end{table}

\subsection{Potential Objections on the Proposed Attacks}
\paragraph{The Flaw was Known.}
That many summarization scoring can be gamed is well known. For example, ROUGE grows when prediction length increases~\cite{sun-etal-2019-compare}. ROUGE-L is not reliable when output space is relatively large~\cite{krishna-etal-2021-hurdles}. That ROUGE correlates badly with human judgments at a system level has been revealed by findings of \citeauthor{paulus2018a}. And, BERTScore does not improve upon the correlation of ROUGE~\cite{fabbri-etal-2021-summeval,gehrmann-etal-2021-gem}.

The current work goes beyond most conventional arguments and analyses against the metrics, and actually constructs a system that sets out to game ROUGE, METEOR, and BERTScore together. We believe that clearly showing the vulnerability is beneficial for scoring remediation efforts. From a behavioural viewpoint, each step of defence against an attack makes the scoring more robust. Compared with findings by \citeauthor{paulus2018a}, we cover more metrics, and provide a more thorough overthrow of the monotonicity of the scoring systems, \emph{i.e.}, outputs from our threat model are significantly worse.

\paragraph{Shoddy Attack?}
The proposed attack is easy to detect, so its effectiveness may be questioned. In fact, since we are the first to see automatic scoring as a decent NLU task and attack the most widely used systems, evasion attacks are relatively easy. This just goes to show that even the crudest attack can work on these scoring systems. Certainly, as the scoring system becomes more robust, the attack has to be more crafted. For example, if the minimum accepted input to the scoring system is a "grammatically correct" sentence, an attacker may have to search for fluent but factually incorrect sentences. With a contest like this, we may end up with a robust scoring system.

As for attack scope, we believe it is more urgent to explore popular metrics, as they currently have the greatest impact on summarization. Nonetheless, we will expand to a wider range of scoring and catch up with emerging ratings such as BLEURT~\cite{sellam-etal-2020-bleurt}.

\subsection{Potential Difficulties}
Performing evasion attacks with bad texts is easy, when texts are as bad as broken sentences or scrambled codes in Table~\ref{tab:example}. In this case, the output of the threat system does not need to be scrutinized by human evaluators. However, human evaluation of attack examples may be required to identify more complex flaws, such as untrue statements or those that the document does not entail. Therefore, more effort may be required when performing evasion attacks on more robust scoring systems.

\section{Conclusion}
We hereby answer the question: it is easy to create a threat system that simultaneously scores high on ROUGE, METEOR, and BERTScore using worse text. In this work, we treat automatic scoring as a regression machine learning task and conduct evasion attacks to probe its robustness or reliability. Our attacker, whose score competes with top-level summarizers, actually outputs non-summary strings. This further suggests that current mainstream scoring systems are not a sufficient condition to support the plausibility of summarizers, as they ignore the linguistic information required to compute sentence proximity. Intentionally or not, optimizing for flawed scores can prevent algorithms from summarizing well. The practical effectiveness of existing summarizing algorithms is not affected by this, since most of them optimize maximum likelihood estimation. Based on the exposed vulnerabilities, careful fixes to scoring systems that measure summary quality and sentence similarity are necessary.

\section*{Ethical considerations}
The techniques developed in this study can be recognized by programs or humans, and we also provide defences. Our intention is not to harm, but to publish such attacks publicly so that better scores can be developed in the future and to better guide the development of summaries. This is similar to how hackers publicly expose bugs/vulnerabilities in software. This shows that our work has long-term benefits for the community. Our attacks are not against real-world machine learning systems.

\section*{Limitations}
We have only attacked the three most widely adopted scoring schemes that have already in summarization literature. However, there are emerging scoring schemes like BLEURT~\cite{sellam-etal-2020-bleurt}, which will be studied in our future work.



\bibliography{anthology,custom}

\begin{thebibliography}{86}
\expandafter\ifx\csname natexlab\endcsname\relax\def\natexlab#1{#1}\fi

\bibitem[{Alzantot et~al.(2018)Alzantot, Sharma, Elgohary, Ho, Srivastava, and
  Chang}]{alzantot-etal-2018-generating}
Moustafa Alzantot, Yash Sharma, Ahmed Elgohary, Bo-Jhang Ho, Mani Srivastava,
  and Kai-Wei Chang. 2018.
\newblock \href {https://doi.org/10.18653/v1/D18-1316} {Generating natural
  language adversarial examples}.
\newblock In \emph{Proceedings of the 2018 Conference on Empirical Methods in
  Natural Language Processing}, pages 2890--2896, Brussels, Belgium.
  Association for Computational Linguistics.

\bibitem[{Bae et~al.(2019)Bae, Kim, Kim, and Lee}]{bae-etal-2019-summary}
Sanghwan Bae, Taeuk Kim, Jihoon Kim, and Sang-goo Lee. 2019.
\newblock \href {https://doi.org/10.18653/v1/D19-5402} {Summary level training
  of sentence rewriting for abstractive summarization}.
\newblock In \emph{Proceedings of the 2nd Workshop on New Frontiers in
  Summarization}, pages 10--20, Hong Kong, China. Association for Computational
  Linguistics.

\bibitem[{Banerjee and Lavie(2005)}]{banerjee-lavie-2005-meteor}
Satanjeev Banerjee and Alon Lavie. 2005.
\newblock \href {https://aclanthology.org/W05-0909} {{METEOR}: An automatic
  metric for {MT} evaluation with improved correlation with human judgments}.
\newblock In \emph{Proceedings of the {ACL} Workshop on Intrinsic and Extrinsic
  Evaluation Measures for Machine Translation and/or Summarization}, pages
  65--72, Ann Arbor, Michigan. Association for Computational Linguistics.

\bibitem[{Behjati et~al.(2019)Behjati, Moosavi-Dezfooli, Baghshah, and
  Frossard}]{behjati2019universal}
Melika Behjati, Seyed-Mohsen Moosavi-Dezfooli, Mahdieh~Soleymani Baghshah, and
  Pascal Frossard. 2019.
\newblock \href {https://doi.org/10.1109/ICASSP.2019.8682430} {Universal
  adversarial attacks on text classifiers}.
\newblock In \emph{ICASSP 2019 - 2019 IEEE International Conference on
  Acoustics, Speech and Signal Processing (ICASSP)}, pages 7345--7349.

\bibitem[{Bertossi et~al.(2018)Bertossi, Gottlob, and
  Pichler}]{bertossi2018datalog}
Leopoldo~E. Bertossi, Georg Gottlob, and Reinhard Pichler. 2018.
\newblock \href {http://arxiv.org/abs/1803.06445} {Datalog: Bag semantics via
  set semantics}.
\newblock \emph{CoRR}, abs/1803.06445.

\bibitem[{Bhandari et~al.(2020)Bhandari, Gour, Ashfaq, and
  Liu}]{bhandari-etal-2020-metrics}
Manik Bhandari, Pranav~Narayan Gour, Atabak Ashfaq, and Pengfei Liu. 2020.
\newblock \href {https://doi.org/10.18653/v1/2020.coling-main.501} {Metrics
  also disagree in the low scoring range: Revisiting summarization evaluation
  metrics}.
\newblock In \emph{Proceedings of the 28th International Conference on
  Computational Linguistics}, pages 5702--5711, Barcelona, Spain (Online).
  International Committee on Computational Linguistics.

\bibitem[{{Blank} and {Deb}(2020)}]{pymoo}
J.~{Blank} and K.~{Deb}. 2020.
\newblock pymoo: Multi-objective optimization in python.
\newblock \emph{IEEE Access}, 8:89497--89509.

\bibitem[{B{\"o}hm et~al.(2019)B{\"o}hm, Gao, Meyer, Shapira, Dagan, and
  Gurevych}]{bohm-etal-2019-better}
Florian B{\"o}hm, Yang Gao, Christian~M. Meyer, Ori Shapira, Ido Dagan, and
  Iryna Gurevych. 2019.
\newblock \href {https://doi.org/10.18653/v1/D19-1307} {Better rewards yield
  better summaries: Learning to summarise without references}.
\newblock In \emph{Proceedings of the 2019 Conference on Empirical Methods in
  Natural Language Processing and the 9th International Joint Conference on
  Natural Language Processing (EMNLP-IJCNLP)}, pages 3110--3120, Hong Kong,
  China. Association for Computational Linguistics.

\bibitem[{Cao and Gong(2017)}]{cao2017mitigating}
Xiaoyu Cao and Neil~Zhenqiang Gong. 2017.
\newblock \href {https://doi.org/10.1145/3134600.3134606} {Mitigating evasion
  attacks to deep neural networks via region-based classification}.
\newblock In \emph{Proceedings of the 33rd Annual Computer Security
  Applications Conference, Orlando, FL, USA, December 4-8, 2017}, pages
  278--287. {ACM}.

\bibitem[{Carlini and Wagner(2017)}]{carlini2017adversarial}
Nicholas Carlini and David Wagner. 2017.
\newblock \href {https://doi.org/10.1145/3128572.3140444} {\emph{Adversarial
  Examples Are Not Easily Detected: Bypassing Ten Detection Methods}}, page
  3–14. Association for Computing Machinery, New York, NY, USA.

\bibitem[{Celikyilmaz et~al.(2018)Celikyilmaz, Bosselut, He, and
  Choi}]{celikyilmaz-etal-2018-deep}
Asli Celikyilmaz, Antoine Bosselut, Xiaodong He, and Yejin Choi. 2018.
\newblock \href {https://doi.org/10.18653/v1/N18-1150} {Deep communicating
  agents for abstractive summarization}.
\newblock In \emph{Proceedings of the 2018 Conference of the North {A}merican
  Chapter of the Association for Computational Linguistics: Human Language
  Technologies, Volume 1 (Long Papers)}, pages 1662--1675, New Orleans,
  Louisiana. Association for Computational Linguistics.

\bibitem[{Cer et~al.(2018)Cer, Yang, Kong, Hua, Limtiaco, St.~John, Constant,
  Guajardo-Cespedes, Yuan, Tar, Strope, and Kurzweil}]{cer-etal-2018-universal}
Daniel Cer, Yinfei Yang, Sheng-yi Kong, Nan Hua, Nicole Limtiaco, Rhomni
  St.~John, Noah Constant, Mario Guajardo-Cespedes, Steve Yuan, Chris Tar,
  Brian Strope, and Ray Kurzweil. 2018.
\newblock \href {https://doi.org/10.18653/v1/D18-2029} {Universal sentence
  encoder for {E}nglish}.
\newblock In \emph{Proceedings of the 2018 Conference on Empirical Methods in
  Natural Language Processing: System Demonstrations}, pages 169--174,
  Brussels, Belgium. Association for Computational Linguistics.

\bibitem[{Chen and Bansal(2018)}]{chen-bansal-2018-fast}
Yen-Chun Chen and Mohit Bansal. 2018.
\newblock \href {https://doi.org/10.18653/v1/P18-1063} {Fast abstractive
  summarization with reinforce-selected sentence rewriting}.
\newblock In \emph{Proceedings of the 56th Annual Meeting of the Association
  for Computational Linguistics (Volume 1: Long Papers)}, pages 675--686,
  Melbourne, Australia. Association for Computational Linguistics.

\bibitem[{Chowdhery et~al.(2022)Chowdhery, Narang, Devlin, Bosma, Mishra,
  Roberts, Barham, Chung, Sutton, Gehrmann, Schuh, Shi, Tsvyashchenko, Maynez,
  Rao, Barnes, Tay, Shazeer, Prabhakaran, Reif, Du, Hutchinson, Pope, Bradbury,
  Austin, Isard, Gur{-}Ari, Yin, Duke, Levskaya, Ghemawat, Dev, Michalewski,
  Garcia, Misra, Robinson, Fedus, Zhou, Ippolito, Luan, Lim, Zoph, Spiridonov,
  Sepassi, Dohan, Agrawal, Omernick, Dai, Pillai, Pellat, Lewkowycz, Moreira,
  Child, Polozov, Lee, Zhou, Wang, Saeta, Diaz, Firat, Catasta, Wei,
  Meier{-}Hellstern, Eck, Dean, Petrov, and Fiedel}]{chowdhery2022palm}
Aakanksha Chowdhery, Sharan Narang, Jacob Devlin, Maarten Bosma, Gaurav Mishra,
  Adam Roberts, Paul Barham, Hyung~Won Chung, Charles Sutton, Sebastian
  Gehrmann, Parker Schuh, Kensen Shi, Sasha Tsvyashchenko, Joshua Maynez,
  Abhishek Rao, Parker Barnes, Yi~Tay, Noam Shazeer, Vinodkumar Prabhakaran,
  Emily Reif, Nan Du, Ben Hutchinson, Reiner Pope, James Bradbury, Jacob
  Austin, Michael Isard, Guy Gur{-}Ari, Pengcheng Yin, Toju Duke, Anselm
  Levskaya, Sanjay Ghemawat, Sunipa Dev, Henryk Michalewski, Xavier Garcia,
  Vedant Misra, Kevin Robinson, Liam Fedus, Denny Zhou, Daphne Ippolito, David
  Luan, Hyeontaek Lim, Barret Zoph, Alexander Spiridonov, Ryan Sepassi, David
  Dohan, Shivani Agrawal, Mark Omernick, Andrew~M. Dai,
  Thanumalayan~Sankaranarayana Pillai, Marie Pellat, Aitor Lewkowycz, Erica
  Moreira, Rewon Child, Oleksandr Polozov, Katherine Lee, Zongwei Zhou, Xuezhi
  Wang, Brennan Saeta, Mark Diaz, Orhan Firat, Michele Catasta, Jason Wei,
  Kathy Meier{-}Hellstern, Douglas Eck, Jeff Dean, Slav Petrov, and Noah
  Fiedel. 2022.
\newblock \href {https://doi.org/10.48550/arXiv.2204.02311} {Palm: Scaling
  language modeling with pathways}.
\newblock \emph{CoRR}, abs/2204.02311.

\bibitem[{Clarke and Lapata(2006)}]{clarke-lapata-2006-models}
James Clarke and Mirella Lapata. 2006.
\newblock \href {https://doi.org/10.3115/1220175.1220223} {Models for sentence
  compression: A comparison across domains, training requirements and
  evaluation measures}.
\newblock In \emph{Proceedings of the 21st International Conference on
  Computational Linguistics and 44th Annual Meeting of the Association for
  Computational Linguistics}, pages 377--384, Sydney, Australia. Association
  for Computational Linguistics.

\bibitem[{Dagan et~al.(2006)Dagan, Glickman, and Magnini}]{dagan2005pascal}
Ido Dagan, Oren Glickman, and Bernardo Magnini. 2006.
\newblock The pascal recognising textual entailment challenge.
\newblock In \emph{Machine Learning Challenges. Evaluating Predictive
  Uncertainty, Visual Object Classification, and Recognising Tectual
  Entailment}, pages 177--190, Berlin, Heidelberg. Springer Berlin Heidelberg.

\bibitem[{Devlin et~al.(2019)Devlin, Chang, Lee, and
  Toutanova}]{devlin-etal-2019-bert}
Jacob Devlin, Ming-Wei Chang, Kenton Lee, and Kristina Toutanova. 2019.
\newblock \href {https://doi.org/10.18653/v1/N19-1423} {{BERT}: Pre-training of
  deep bidirectional transformers for language understanding}.
\newblock In \emph{Proceedings of the 2019 Conference of the North {A}merican
  Chapter of the Association for Computational Linguistics: Human Language
  Technologies, Volume 1 (Long and Short Papers)}, pages 4171--4186,
  Minneapolis, Minnesota. Association for Computational Linguistics.

\bibitem[{Dong et~al.(2019)Dong, Yang, Wang, Wei, Liu, Wang, Gao, Zhou, and
  Hon}]{dong2019unified}
Li~Dong, Nan Yang, Wenhui Wang, Furu Wei, Xiaodong Liu, Yu~Wang, Jianfeng Gao,
  Ming Zhou, and Hsiao{-}Wuen Hon. 2019.
\newblock \href
  {https://proceedings.neurips.cc/paper/2019/hash/c20bb2d9a50d5ac1f713f8b34d9aac5a-Abstract.html}
  {Unified language model pre-training for natural language understanding and
  generation}.
\newblock In \emph{Advances in Neural Information Processing Systems 32: Annual
  Conference on Neural Information Processing Systems 2019, NeurIPS 2019,
  December 8-14, 2019, Vancouver, BC, Canada}, pages 13042--13054.

\bibitem[{Dong et~al.(2018)Dong, Shen, Crawford, van Hoof, and
  Cheung}]{dong-etal-2018-banditsum}
Yue Dong, Yikang Shen, Eric Crawford, Herke van Hoof, and Jackie Chi~Kit
  Cheung. 2018.
\newblock \href {https://doi.org/10.18653/v1/D18-1409} {{B}andit{S}um:
  Extractive summarization as a contextual bandit}.
\newblock In \emph{Proceedings of the 2018 Conference on Empirical Methods in
  Natural Language Processing}, pages 3739--3748, Brussels, Belgium.
  Association for Computational Linguistics.

\bibitem[{Dou et~al.(2021)Dou, Liu, Hayashi, Jiang, and
  Neubig}]{dou-etal-2021-gsum}
Zi-Yi Dou, Pengfei Liu, Hiroaki Hayashi, Zhengbao Jiang, and Graham Neubig.
  2021.
\newblock \href {https://doi.org/10.18653/v1/2021.naacl-main.384} {{GS}um: A
  general framework for guided neural abstractive summarization}.
\newblock In \emph{Proceedings of the 2021 Conference of the North American
  Chapter of the Association for Computational Linguistics: Human Language
  Technologies}, pages 4830--4842, Online. Association for Computational
  Linguistics.

\bibitem[{Fabbri et~al.(2021)Fabbri, Kry{\'s}ci{\'n}ski, McCann, Xiong, Socher,
  and Radev}]{fabbri-etal-2021-summeval}
Alexander~R. Fabbri, Wojciech Kry{\'s}ci{\'n}ski, Bryan McCann, Caiming Xiong,
  Richard Socher, and Dragomir Radev. 2021.
\newblock \href {https://doi.org/10.1162/tacl_a_00373} {{S}umm{E}val:
  Re-evaluating summarization evaluation}.
\newblock \emph{Transactions of the Association for Computational Linguistics},
  9:391--409.

\bibitem[{Filighera et~al.(2022)Filighera, Ochs, Steuer, and
  Tregel}]{filighera2022cheating}
Anna Filighera, Sebastian Ochs, Tim Steuer, and Thomas Tregel. 2022.
\newblock \href {http://arxiv.org/abs/2201.08318} {Cheating automatic short
  answer grading: On the adversarial usage of adjectives and adverbs}.
\newblock \emph{CoRR}, abs/2201.08318.

\bibitem[{Filighera et~al.(2020)Filighera, Steuer, and
  Rensing}]{filighera2020fooling}
Anna Filighera, Tim Steuer, and Christoph Rensing. 2020.
\newblock Fooling automatic short answer grading systems.
\newblock In \emph{Artificial Intelligence in Education}, pages 177--190, Cham.
  Springer International Publishing.

\bibitem[{Gehrmann et~al.(2021)Gehrmann, Adewumi, Aggarwal, Ammanamanchi,
  Aremu, Bosselut, Chandu, Clinciu, Das, Dhole, Du, Durmus, Du{\v{s}}ek,
  Emezue, Gangal, Garbacea, Hashimoto, Hou, Jernite, Jhamtani, Ji, Jolly, Kale,
  Kumar, Ladhak, Madaan, Maddela, Mahajan, Mahamood, Majumder, Martins,
  McMillan-Major, Mille, van Miltenburg, Nadeem, Narayan, Nikolaev,
  Niyongabo~Rubungo, Osei, Parikh, Perez-Beltrachini, Rao, Raunak, Rodriguez,
  Santhanam, Sedoc, Sellam, Shaikh, Shimorina, Sobrevilla~Cabezudo, Strobelt,
  Subramani, Xu, Yang, Yerukola, and Zhou}]{gehrmann-etal-2021-gem}
Sebastian Gehrmann, Tosin Adewumi, Karmanya Aggarwal, Pawan~Sasanka
  Ammanamanchi, Anuoluwapo Aremu, Antoine Bosselut, Khyathi~Raghavi Chandu,
  Miruna-Adriana Clinciu, Dipanjan Das, Kaustubh Dhole, Wanyu Du, Esin Durmus,
  Ond{\v{r}}ej Du{\v{s}}ek, Chris~Chinenye Emezue, Varun Gangal, Cristina
  Garbacea, Tatsunori Hashimoto, Yufang Hou, Yacine Jernite, Harsh Jhamtani,
  Yangfeng Ji, Shailza Jolly, Mihir Kale, Dhruv Kumar, Faisal Ladhak, Aman
  Madaan, Mounica Maddela, Khyati Mahajan, Saad Mahamood, Bodhisattwa~Prasad
  Majumder, Pedro~Henrique Martins, Angelina McMillan-Major, Simon Mille, Emiel
  van Miltenburg, Moin Nadeem, Shashi Narayan, Vitaly Nikolaev, Andre
  Niyongabo~Rubungo, Salomey Osei, Ankur Parikh, Laura Perez-Beltrachini,
  Niranjan~Ramesh Rao, Vikas Raunak, Juan~Diego Rodriguez, Sashank Santhanam,
  Jo{\~a}o Sedoc, Thibault Sellam, Samira Shaikh, Anastasia Shimorina,
  Marco~Antonio Sobrevilla~Cabezudo, Hendrik Strobelt, Nishant Subramani, Wei
  Xu, Diyi Yang, Akhila Yerukola, and Jiawei Zhou. 2021.
\newblock \href {https://doi.org/10.18653/v1/2021.gem-1.10} {The {GEM}
  benchmark: Natural language generation, its evaluation and metrics}.
\newblock In \emph{Proceedings of the 1st Workshop on Natural Language
  Generation, Evaluation, and Metrics (GEM 2021)}, pages 96--120, Online.
  Association for Computational Linguistics.

\bibitem[{Gehrmann et~al.(2018)Gehrmann, Deng, and
  Rush}]{gehrmann-etal-2018-bottom}
Sebastian Gehrmann, Yuntian Deng, and Alexander Rush. 2018.
\newblock \href {https://doi.org/10.18653/v1/D18-1443} {Bottom-up abstractive
  summarization}.
\newblock In \emph{Proceedings of the 2018 Conference on Empirical Methods in
  Natural Language Processing}, pages 4098--4109, Brussels, Belgium.
  Association for Computational Linguistics.

\bibitem[{Geirhos et~al.(2020)Geirhos, Jacobsen, Michaelis, Zemel, Brendel,
  Bethge, and Wichmann}]{geirhos2020shortcut}
Robert Geirhos, J{\"{o}}rn{-}Henrik Jacobsen, Claudio Michaelis, Richard~S.
  Zemel, Wieland Brendel, Matthias Bethge, and Felix~A. Wichmann. 2020.
\newblock \href {https://doi.org/10.1038/s42256-020-00257-z} {Shortcut learning
  in deep neural networks}.
\newblock \emph{Nat. Mach. Intell.}, 2(11):665--673.

\bibitem[{Graham(2015)}]{graham-2015-evaluating}
Yvette Graham. 2015.
\newblock \href {https://doi.org/10.18653/v1/D15-1013} {Re-evaluating automatic
  summarization with {BLEU} and 192 shades of {ROUGE}}.
\newblock In \emph{Proceedings of the 2015 Conference on Empirical Methods in
  Natural Language Processing}, pages 128--137, Lisbon, Portugal. Association
  for Computational Linguistics.

\bibitem[{Guo et~al.(2018)Guo, Pasunuru, and Bansal}]{guo-etal-2018-soft}
Han Guo, Ramakanth Pasunuru, and Mohit Bansal. 2018.
\newblock \href {https://doi.org/10.18653/v1/P18-1064} {Soft layer-specific
  multi-task summarization with entailment and question generation}.
\newblock In \emph{Proceedings of the 56th Annual Meeting of the Association
  for Computational Linguistics (Volume 1: Long Papers)}, pages 687--697,
  Melbourne, Australia. Association for Computational Linguistics.

\bibitem[{Holland(2012)}]{holland1992genetic}
John~H. Holland. 2012.
\newblock \href {https://doi.org/10.4249/scholarpedia.1482} {Genetic
  algorithms}.
\newblock \emph{Scholarpedia}, 7(12):1482.

\bibitem[{Hsu et~al.(2018)Hsu, Lin, Lee, Min, Tang, and
  Sun}]{hsu-etal-2018-unified}
Wan-Ting Hsu, Chieh-Kai Lin, Ming-Ying Lee, Kerui Min, Jing Tang, and Min Sun.
  2018.
\newblock \href {https://doi.org/10.18653/v1/P18-1013} {A unified model for
  extractive and abstractive summarization using inconsistency loss}.
\newblock In \emph{Proceedings of the 56th Annual Meeting of the Association
  for Computational Linguistics (Volume 1: Long Papers)}, pages 132--141,
  Melbourne, Australia. Association for Computational Linguistics.

\bibitem[{Jauregi~Unanue et~al.(2021)Jauregi~Unanue, Parnell, and
  Piccardi}]{jauregi-unanue-etal-2021-berttune}
Inigo Jauregi~Unanue, Jacob Parnell, and Massimo Piccardi. 2021.
\newblock \href {https://doi.org/10.18653/v1/2021.acl-short.115} {{BERTT}une:
  Fine-tuning neural machine translation with {BERTS}core}.
\newblock In \emph{Proceedings of the 59th Annual Meeting of the Association
  for Computational Linguistics and the 11th International Joint Conference on
  Natural Language Processing (Volume 2: Short Papers)}, pages 915--924,
  Online. Association for Computational Linguistics.

\bibitem[{Jiang and Bansal(2018)}]{jiang-bansal-2018-closed}
Yichen Jiang and Mohit Bansal. 2018.
\newblock \href {https://doi.org/10.18653/v1/D18-1440} {Closed-book training to
  improve summarization encoder memory}.
\newblock In \emph{Proceedings of the 2018 Conference on Empirical Methods in
  Natural Language Processing}, pages 4067--4077, Brussels, Belgium.
  Association for Computational Linguistics.

\bibitem[{Ke and Ng(2019)}]{ke2019automated}
Zixuan Ke and Vincent Ng. 2019.
\newblock \href {https://doi.org/10.24963/ijcai.2019/879} {Automated essay
  scoring: A survey of the state of the art}.
\newblock In \emph{Proceedings of the Twenty-Eighth International Joint
  Conference on Artificial Intelligence, {IJCAI-19}}, pages 6300--6308.
  International Joint Conferences on Artificial Intelligence Organization.

\bibitem[{Krishna et~al.(2021)Krishna, Roy, and
  Iyyer}]{krishna-etal-2021-hurdles}
Kalpesh Krishna, Aurko Roy, and Mohit Iyyer. 2021.
\newblock \href {https://doi.org/10.18653/v1/2021.naacl-main.393} {Hurdles to
  progress in long-form question answering}.
\newblock In \emph{Proceedings of the 2021 Conference of the North American
  Chapter of the Association for Computational Linguistics: Human Language
  Technologies}, pages 4940--4957, Online. Association for Computational
  Linguistics.

\bibitem[{Kryscinski et~al.(2019)Kryscinski, Keskar, McCann, Xiong, and
  Socher}]{kryscinski-etal-2019-neural}
Wojciech Kryscinski, Nitish~Shirish Keskar, Bryan McCann, Caiming Xiong, and
  Richard Socher. 2019.
\newblock \href {https://doi.org/10.18653/v1/D19-1051} {Neural text
  summarization: A critical evaluation}.
\newblock In \emph{Proceedings of the 2019 Conference on Empirical Methods in
  Natural Language Processing and the 9th International Joint Conference on
  Natural Language Processing (EMNLP-IJCNLP)}, pages 540--551, Hong Kong,
  China. Association for Computational Linguistics.

\bibitem[{Kry{\'s}ci{\'n}ski et~al.(2018)Kry{\'s}ci{\'n}ski, Paulus, Xiong, and
  Socher}]{kryscinski-etal-2018-improving}
Wojciech Kry{\'s}ci{\'n}ski, Romain Paulus, Caiming Xiong, and Richard Socher.
  2018.
\newblock \href {https://doi.org/10.18653/v1/D18-1207} {Improving abstraction
  in text summarization}.
\newblock In \emph{Proceedings of the 2018 Conference on Empirical Methods in
  Natural Language Processing}, pages 1808--1817, Brussels, Belgium.
  Association for Computational Linguistics.

\bibitem[{Lamontagne and Abi-Zeid(2006)}]{lamontagne2006combining}
Luc Lamontagne and Ir{\`e}ne Abi-Zeid. 2006.
\newblock Combining multiple similarity metrics using a multicriteria approach.
\newblock In \emph{Advances in Case-Based Reasoning}, pages 415--428, Berlin,
  Heidelberg. Springer Berlin Heidelberg.

\bibitem[{Lavie and Agarwal(2007)}]{lavie-agarwal-2007-meteor}
Alon Lavie and Abhaya Agarwal. 2007.
\newblock \href {https://aclanthology.org/W07-0734} {{METEOR}: An automatic
  metric for {MT} evaluation with high levels of correlation with human
  judgments}.
\newblock In \emph{Proceedings of the Second Workshop on Statistical Machine
  Translation}, pages 228--231, Prague, Czech Republic. Association for
  Computational Linguistics.

\bibitem[{Lewis et~al.(2020)Lewis, Liu, Goyal, Ghazvininejad, Mohamed, Levy,
  Stoyanov, and Zettlemoyer}]{lewis-etal-2020-bart}
Mike Lewis, Yinhan Liu, Naman Goyal, Marjan Ghazvininejad, Abdelrahman Mohamed,
  Omer Levy, Veselin Stoyanov, and Luke Zettlemoyer. 2020.
\newblock \href {https://doi.org/10.18653/v1/2020.acl-main.703} {{BART}:
  Denoising sequence-to-sequence pre-training for natural language generation,
  translation, and comprehension}.
\newblock In \emph{Proceedings of the 58th Annual Meeting of the Association
  for Computational Linguistics}, pages 7871--7880, Online. Association for
  Computational Linguistics.

\bibitem[{Li et~al.(2018{\natexlab{a}})Li, Xiao, Lyu, and
  Wang}]{li-etal-2018-improving}
Wei Li, Xinyan Xiao, Yajuan Lyu, and Yuanzhuo Wang. 2018{\natexlab{a}}.
\newblock \href {https://doi.org/10.18653/v1/D18-1205} {Improving neural
  abstractive document summarization with explicit information selection
  modeling}.
\newblock In \emph{Proceedings of the 2018 Conference on Empirical Methods in
  Natural Language Processing}, pages 1787--1796, Brussels, Belgium.
  Association for Computational Linguistics.

\bibitem[{Li et~al.(2018{\natexlab{b}})Li, Xiao, Lyu, and
  Wang}]{li-etal-2018-improving-neural}
Wei Li, Xinyan Xiao, Yajuan Lyu, and Yuanzhuo Wang. 2018{\natexlab{b}}.
\newblock \href {https://doi.org/10.18653/v1/D18-1441} {Improving neural
  abstractive document summarization with structural regularization}.
\newblock In \emph{Proceedings of the 2018 Conference on Empirical Methods in
  Natural Language Processing}, pages 4078--4087, Brussels, Belgium.
  Association for Computational Linguistics.

\bibitem[{Lin(2004)}]{lin-2004-rouge}
Chin-Yew Lin. 2004.
\newblock \href {https://aclanthology.org/W04-1013} {{ROUGE}: A package for
  automatic evaluation of summaries}.
\newblock In \emph{Text Summarization Branches Out}, pages 74--81, Barcelona,
  Spain. Association for Computational Linguistics.

\bibitem[{Lin and Hovy(2003)}]{lin-hovy-2003-automatic}
Chin-Yew Lin and Eduard Hovy. 2003.
\newblock \href {https://aclanthology.org/N03-1020} {Automatic evaluation of
  summaries using n-gram co-occurrence statistics}.
\newblock In \emph{Proceedings of the 2003 Human Language Technology Conference
  of the North {A}merican Chapter of the Association for Computational
  Linguistics}, pages 150--157.

\bibitem[{Liu and Lapata(2019)}]{liu-lapata-2019-text}
Yang Liu and Mirella Lapata. 2019.
\newblock \href {https://doi.org/10.18653/v1/D19-1387} {Text summarization with
  pretrained encoders}.
\newblock In \emph{Proceedings of the 2019 Conference on Empirical Methods in
  Natural Language Processing and the 9th International Joint Conference on
  Natural Language Processing (EMNLP-IJCNLP)}, pages 3730--3740, Hong Kong,
  China. Association for Computational Linguistics.

\bibitem[{Liu et~al.(2019)Liu, Ott, Goyal, Du, Joshi, Chen, Levy, Lewis,
  Zettlemoyer, and Stoyanov}]{liu2019roberta}
Yinhan Liu, Myle Ott, Naman Goyal, Jingfei Du, Mandar Joshi, Danqi Chen, Omer
  Levy, Mike Lewis, Luke Zettlemoyer, and Veselin Stoyanov. 2019.
\newblock \href {http://arxiv.org/abs/1907.11692} {Roberta: {A} robustly
  optimized {BERT} pretraining approach}.
\newblock \emph{CoRR}, abs/1907.11692.

\bibitem[{Liu and Liu(2021)}]{liu-liu-2021-simcls}
Yixin Liu and Pengfei Liu. 2021.
\newblock \href {https://doi.org/10.18653/v1/2021.acl-short.135} {{S}im{CLS}: A
  simple framework for contrastive learning of abstractive summarization}.
\newblock In \emph{Proceedings of the 59th Annual Meeting of the Association
  for Computational Linguistics and the 11th International Joint Conference on
  Natural Language Processing (Volume 2: Short Papers)}, pages 1065--1072,
  Online. Association for Computational Linguistics.

\bibitem[{Maheshwary et~al.(2021)Maheshwary, Maheshwary, and
  Pudi}]{maheshwary2021generating}
Rishabh Maheshwary, Saket Maheshwary, and Vikram Pudi. 2021.
\newblock \href {https://ojs.aaai.org/index.php/AAAI/article/view/17595}
  {Generating natural language attacks in a hard label black box setting}.
\newblock In \emph{Thirty-Fifth {AAAI} Conference on Artificial Intelligence,
  {AAAI} 2021, Thirty-Third Conference on Innovative Applications of Artificial
  Intelligence, {IAAI} 2021, The Eleventh Symposium on Educational Advances in
  Artificial Intelligence, {EAAI} 2021, Virtual Event, February 2-9, 2021},
  pages 13525--13533. {AAAI} Press.

\bibitem[{Metzen et~al.(2017)Metzen, Genewein, Fischer, and
  Bischoff}]{metzen2017detecting}
Jan~Hendrik Metzen, Tim Genewein, Volker Fischer, and Bastian Bischoff. 2017.
\newblock \href {https://openreview.net/forum?id=SJzCSf9xg} {On detecting
  adversarial perturbations}.
\newblock In \emph{5th International Conference on Learning Representations,
  {ICLR} 2017, Toulon, France, April 24-26, 2017, Conference Track
  Proceedings}. OpenReview.net.

\bibitem[{Moroshko et~al.(2019)Moroshko, Feigenblat, Roitman, and
  Konopnicki}]{moroshko-etal-2019-editorial}
Edward Moroshko, Guy Feigenblat, Haggai Roitman, and David Konopnicki. 2019.
\newblock \href {https://doi.org/10.18653/v1/D19-5407} {An editorial network
  for enhanced document summarization}.
\newblock In \emph{Proceedings of the 2nd Workshop on New Frontiers in
  Summarization}, pages 57--63, Hong Kong, China. Association for Computational
  Linguistics.

\bibitem[{Morris(2020)}]{morris-2020-second}
John Morris. 2020.
\newblock \href {https://doi.org/10.18653/v1/2020.blackboxnlp-1.22}
  {Second-order {NLP} adversarial examples}.
\newblock In \emph{Proceedings of the Third BlackboxNLP Workshop on Analyzing
  and Interpreting Neural Networks for NLP}, pages 228--237, Online.
  Association for Computational Linguistics.

\bibitem[{Nallapati et~al.(2017)Nallapati, Zhai, and
  Zhou}]{nallapati2017summarunner}
Ramesh Nallapati, Feifei Zhai, and Bowen Zhou. 2017.
\newblock \href {http://aaai.org/ocs/index.php/AAAI/AAAI17/paper/view/14636}
  {Summarunner: {A} recurrent neural network based sequence model for
  extractive summarization of documents}.
\newblock In \emph{Proceedings of the Thirty-First {AAAI} Conference on
  Artificial Intelligence, February 4-9, 2017, San Francisco, California,
  {USA}}, pages 3075--3081. {AAAI} Press.

\bibitem[{Nallapati et~al.(2016)Nallapati, Zhou, Gulcehre, Xiang
  et~al.}]{nallapati2016abstractive}
Ramesh Nallapati, Bowen Zhou, Caglar Gulcehre, Bing Xiang, et~al. 2016.
\newblock Abstractive text summarization using sequence-to-sequence rnns and
  beyond.
\newblock \emph{arXiv preprint arXiv:1602.06023}.

\bibitem[{Napoles et~al.(2016)Napoles, Sakaguchi, and
  Tetreault}]{napoles-etal-2016-theres}
Courtney Napoles, Keisuke Sakaguchi, and Joel Tetreault. 2016.
\newblock \href {https://doi.org/10.18653/v1/D16-1228} {There{'}s no
  comparison: Reference-less evaluation metrics in grammatical error
  correction}.
\newblock In \emph{Proceedings of the 2016 Conference on Empirical Methods in
  Natural Language Processing}, pages 2109--2115, Austin, Texas. Association
  for Computational Linguistics.

\bibitem[{Narayan et~al.(2018)Narayan, Cohen, and
  Lapata}]{narayan-etal-2018-ranking}
Shashi Narayan, Shay~B. Cohen, and Mirella Lapata. 2018.
\newblock \href {https://doi.org/10.18653/v1/N18-1158} {Ranking sentences for
  extractive summarization with reinforcement learning}.
\newblock In \emph{Proceedings of the 2018 Conference of the North {A}merican
  Chapter of the Association for Computational Linguistics: Human Language
  Technologies, Volume 1 (Long Papers)}, pages 1747--1759, New Orleans,
  Louisiana. Association for Computational Linguistics.

\bibitem[{Novikova et~al.(2017)Novikova, Du{\v{s}}ek, Cercas~Curry, and
  Rieser}]{novikova-etal-2017-need}
Jekaterina Novikova, Ond{\v{r}}ej Du{\v{s}}ek, Amanda Cercas~Curry, and Verena
  Rieser. 2017.
\newblock \href {https://doi.org/10.18653/v1/D17-1238} {Why we need new
  evaluation metrics for {NLG}}.
\newblock In \emph{Proceedings of the 2017 Conference on Empirical Methods in
  Natural Language Processing}, pages 2241--2252, Copenhagen, Denmark.
  Association for Computational Linguistics.

\bibitem[{Papineni et~al.(2002)Papineni, Roukos, Ward, and
  Zhu}]{papineni-etal-2002-bleu}
Kishore Papineni, Salim Roukos, Todd Ward, and Wei-Jing Zhu. 2002.
\newblock \href {https://doi.org/10.3115/1073083.1073135} {{B}leu: a method for
  automatic evaluation of machine translation}.
\newblock In \emph{Proceedings of the 40th Annual Meeting of the Association
  for Computational Linguistics}, pages 311--318, Philadelphia, Pennsylvania,
  USA. Association for Computational Linguistics.

\bibitem[{Pasunuru and Bansal(2018)}]{pasunuru-bansal-2018-multi}
Ramakanth Pasunuru and Mohit Bansal. 2018.
\newblock \href {https://doi.org/10.18653/v1/N18-2102} {Multi-reward reinforced
  summarization with saliency and entailment}.
\newblock In \emph{Proceedings of the 2018 Conference of the North {A}merican
  Chapter of the Association for Computational Linguistics: Human Language
  Technologies, Volume 2 (Short Papers)}, pages 646--653, New Orleans,
  Louisiana. Association for Computational Linguistics.

\bibitem[{Paulus et~al.(2018)Paulus, Xiong, and Socher}]{paulus2018a}
Romain Paulus, Caiming Xiong, and Richard Socher. 2018.
\newblock \href {https://openreview.net/forum?id=HkAClQgA-} {A deep reinforced
  model for abstractive summarization}.
\newblock In \emph{International Conference on Learning Representations}.

\bibitem[{Peyrard et~al.(2017)Peyrard, Botschen, and
  Gurevych}]{peyrard-etal-2017-learning}
Maxime Peyrard, Teresa Botschen, and Iryna Gurevych. 2017.
\newblock \href {https://doi.org/10.18653/v1/W17-4510} {Learning to score
  system summaries for better content selection evaluation.}
\newblock In \emph{Proceedings of the Workshop on New Frontiers in
  Summarization}, pages 74--84, Copenhagen, Denmark. Association for
  Computational Linguistics.

\bibitem[{Qi et~al.(2020)Qi, Yan, Gong, Liu, Duan, Chen, Zhang, and
  Zhou}]{qi-etal-2020-prophetnet}
Weizhen Qi, Yu~Yan, Yeyun Gong, Dayiheng Liu, Nan Duan, Jiusheng Chen, Ruofei
  Zhang, and Ming Zhou. 2020.
\newblock \href {https://doi.org/10.18653/v1/2020.findings-emnlp.217}
  {{P}rophet{N}et: Predicting future n-gram for
  sequence-to-{S}equence{P}re-training}.
\newblock In \emph{Findings of the Association for Computational Linguistics:
  EMNLP 2020}, pages 2401--2410, Online. Association for Computational
  Linguistics.

\bibitem[{Raffel et~al.(2019)Raffel, Shazeer, Roberts, Lee, Narang, Matena,
  Zhou, Li, and Liu}]{raffel2019exploring}
Colin Raffel, Noam Shazeer, Adam Roberts, Katherine Lee, Sharan Narang, Michael
  Matena, Yanqi Zhou, Wei Li, and Peter~J. Liu. 2019.
\newblock \href {http://arxiv.org/abs/1910.10683} {Exploring the limits of
  transfer learning with a unified text-to-text transformer}.
\newblock \emph{CoRR}, abs/1910.10683.

\bibitem[{Rankel et~al.(2013)Rankel, Conroy, Dang, and
  Nenkova}]{rankel-etal-2013-decade}
Peter~A. Rankel, John~M. Conroy, Hoa~Trang Dang, and Ani Nenkova. 2013.
\newblock \href {https://aclanthology.org/P13-2024} {A decade of automatic
  content evaluation of news summaries: Reassessing the state of the art}.
\newblock In \emph{Proceedings of the 51st Annual Meeting of the Association
  for Computational Linguistics (Volume 2: Short Papers)}, pages 131--136,
  Sofia, Bulgaria. Association for Computational Linguistics.

\bibitem[{Reiter(2018)}]{reiter-2018-structured}
Ehud Reiter. 2018.
\newblock \href {https://doi.org/10.1162/coli_a_00322} {A structured review of
  the validity of {BLEU}}.
\newblock \emph{Computational Linguistics}, 44(3):393--401.

\bibitem[{Riezler et~al.(2003)Riezler, King, Crouch, and
  Zaenen}]{riezler-etal-2003-statistical}
Stefan Riezler, Tracy~H. King, Richard Crouch, and Annie Zaenen. 2003.
\newblock \href {https://aclanthology.org/N03-1026} {Statistical sentence
  condensation using ambiguity packing and stochastic disambiguation methods
  for {L}exical-{F}unctional {G}rammar}.
\newblock In \emph{Proceedings of the 2003 Human Language Technology Conference
  of the North {A}merican Chapter of the Association for Computational
  Linguistics}, pages 197--204.

\bibitem[{Sai et~al.(2022)Sai, Mohankumar, and Khapra}]{sai2022survey}
Ananya~B. Sai, Akash~Kumar Mohankumar, and Mitesh~M. Khapra. 2022.
\newblock \href {https://doi.org/10.1145/3485766} {A survey of evaluation
  metrics used for nlg systems}.
\newblock \emph{ACM Comput. Surv.}, 55(2).

\bibitem[{Schluter(2017)}]{schluter-2017-limits}
Natalie Schluter. 2017.
\newblock \href {https://aclanthology.org/E17-2007} {The limits of automatic
  summarisation according to {ROUGE}}.
\newblock In \emph{Proceedings of the 15th Conference of the {E}uropean Chapter
  of the Association for Computational Linguistics: Volume 2, Short Papers},
  pages 41--45, Valencia, Spain. Association for Computational Linguistics.

\bibitem[{See et~al.(2017)See, Liu, and Manning}]{see-etal-2017-get}
Abigail See, Peter~J. Liu, and Christopher~D. Manning. 2017.
\newblock \href {https://doi.org/10.18653/v1/P17-1099} {Get to the point:
  Summarization with pointer-generator networks}.
\newblock In \emph{Proceedings of the 55th Annual Meeting of the Association
  for Computational Linguistics (Volume 1: Long Papers)}, pages 1073--1083,
  Vancouver, Canada. Association for Computational Linguistics.

\bibitem[{Sellam et~al.(2020)Sellam, Das, and Parikh}]{sellam-etal-2020-bleurt}
Thibault Sellam, Dipanjan Das, and Ankur Parikh. 2020.
\newblock \href {https://doi.org/10.18653/v1/2020.acl-main.704} {{BLEURT}:
  Learning robust metrics for text generation}.
\newblock In \emph{Proceedings of the 58th Annual Meeting of the Association
  for Computational Linguistics}, pages 7881--7892, Online. Association for
  Computational Linguistics.

\bibitem[{Song et~al.(2021)Song, Yu, Peng, and
  Narasimhan}]{song-etal-2021-universal}
Liwei Song, Xinwei Yu, Hsuan-Tung Peng, and Karthik Narasimhan. 2021.
\newblock \href {https://doi.org/10.18653/v1/2021.naacl-main.291} {Universal
  adversarial attacks with natural triggers for text classification}.
\newblock In \emph{Proceedings of the 2021 Conference of the North American
  Chapter of the Association for Computational Linguistics: Human Language
  Technologies}, pages 3724--3733, Online. Association for Computational
  Linguistics.

\bibitem[{Sun et~al.(2019)Sun, Shapira, Dagan, and
  Nenkova}]{sun-etal-2019-compare}
Simeng Sun, Ori Shapira, Ido Dagan, and Ani Nenkova. 2019.
\newblock \href {https://doi.org/10.18653/v1/W19-2303} {How to compare
  summarizers without target length? pitfalls, solutions and re-examination of
  the neural summarization literature}.
\newblock In \emph{Proceedings of the Workshop on Methods for Optimizing and
  Evaluating Neural Language Generation}, pages 21--29, Minneapolis, Minnesota.
  Association for Computational Linguistics.

\bibitem[{Szegedy et~al.(2014)Szegedy, Zaremba, Sutskever, Bruna, Erhan,
  Goodfellow, and Fergus}]{szegedy2013intriguing}
Christian Szegedy, Wojciech Zaremba, Ilya Sutskever, Joan Bruna, Dumitru Erhan,
  Ian~J. Goodfellow, and Rob Fergus. 2014.
\newblock \href {http://arxiv.org/abs/1312.6199} {Intriguing properties of
  neural networks}.
\newblock In \emph{2nd International Conference on Learning Representations,
  {ICLR} 2014, Banff, AB, Canada, April 14-16, 2014, Conference Track
  Proceedings}.

\bibitem[{{van der Lee} et~al.(2021){van der Lee}, Gatt, {van Miltenburg}, and
  Krahmer}]{van2021human}
Chris {van der Lee}, Albert Gatt, Emiel {van Miltenburg}, and Emiel Krahmer.
  2021.
\newblock \href {https://doi.org/https://doi.org/10.1016/j.csl.2020.101151}
  {Human evaluation of automatically generated text: Current trends and best
  practice guidelines}.
\newblock \emph{Computer Speech \& Language}, 67:101151.

\bibitem[{Wallace et~al.(2019)Wallace, Feng, Kandpal, Gardner, and
  Singh}]{wallace-etal-2019-universal}
Eric Wallace, Shi Feng, Nikhil Kandpal, Matt Gardner, and Sameer Singh. 2019.
\newblock \href {https://doi.org/10.18653/v1/D19-1221} {Universal adversarial
  triggers for attacking and analyzing {NLP}}.
\newblock In \emph{Proceedings of the 2019 Conference on Empirical Methods in
  Natural Language Processing and the 9th International Joint Conference on
  Natural Language Processing (EMNLP-IJCNLP)}, pages 2153--2162, Hong Kong,
  China. Association for Computational Linguistics.

\bibitem[{Wang et~al.(2020)Wang, Liu, Zheng, Qiu, and
  Huang}]{wang-etal-2020-heterogeneous}
Danqing Wang, Pengfei Liu, Yining Zheng, Xipeng Qiu, and Xuanjing Huang. 2020.
\newblock \href {https://doi.org/10.18653/v1/2020.acl-main.553} {Heterogeneous
  graph neural networks for extractive document summarization}.
\newblock In \emph{Proceedings of the 58th Annual Meeting of the Association
  for Computational Linguistics}, pages 6209--6219, Online. Association for
  Computational Linguistics.

\bibitem[{Xu et~al.(2020)Xu, Gan, Cheng, and Liu}]{xu-etal-2020-discourse}
Jiacheng Xu, Zhe Gan, Yu~Cheng, and Jingjing Liu. 2020.
\newblock \href {https://doi.org/10.18653/v1/2020.acl-main.451}
  {Discourse-aware neural extractive text summarization}.
\newblock In \emph{Proceedings of the 58th Annual Meeting of the Association
  for Computational Linguistics}, pages 5021--5031, Online. Association for
  Computational Linguistics.

\bibitem[{Xu et~al.(2018)Xu, Evans, and Qi}]{xu2017feature}
Weilin Xu, David Evans, and Yanjun Qi. 2018.
\newblock \href
  {http://wp.internetsociety.org/ndss/wp-content/uploads/sites/25/2018/02/ndss2018\_03A-4\_Xu\_paper.pdf}
  {Feature squeezing: Detecting adversarial examples in deep neural networks}.
\newblock In \emph{25th Annual Network and Distributed System Security
  Symposium, {NDSS} 2018, San Diego, California, USA, February 18-21, 2018}.
  The Internet Society.

\bibitem[{Zang et~al.(2020)Zang, Qi, Yang, Liu, Zhang, Liu, and
  Sun}]{zang-etal-2020-word}
Yuan Zang, Fanchao Qi, Chenghao Yang, Zhiyuan Liu, Meng Zhang, Qun Liu, and
  Maosong Sun. 2020.
\newblock \href {https://doi.org/10.18653/v1/2020.acl-main.540} {Word-level
  textual adversarial attacking as combinatorial optimization}.
\newblock In \emph{Proceedings of the 58th Annual Meeting of the Association
  for Computational Linguistics}, pages 6066--6080, Online. Association for
  Computational Linguistics.

\bibitem[{Zhang et~al.(2019{\natexlab{a}})Zhang, Cai, Xu, and
  Wang}]{zhang-etal-2019-pretraining}
Haoyu Zhang, Jingjing Cai, Jianjun Xu, and Ji~Wang. 2019{\natexlab{a}}.
\newblock \href {https://doi.org/10.18653/v1/K19-1074} {Pretraining-based
  natural language generation for text summarization}.
\newblock In \emph{Proceedings of the 23rd Conference on Computational Natural
  Language Learning (CoNLL)}, pages 789--797, Hong Kong, China. Association for
  Computational Linguistics.

\bibitem[{Zhang et~al.(2020)Zhang, Zhao, Saleh, and Liu}]{zhang2020pegasus}
Jingqing Zhang, Yao Zhao, Mohammad Saleh, and Peter~J. Liu. 2020.
\newblock \href {http://proceedings.mlr.press/v119/zhang20ae.html} {{PEGASUS:}
  pre-training with extracted gap-sentences for abstractive summarization}.
\newblock In \emph{Proceedings of the 37th International Conference on Machine
  Learning, {ICML} 2020, 13-18 July 2020, Virtual Event}, volume 119 of
  \emph{Proceedings of Machine Learning Research}, pages 11328--11339. {PMLR}.

\bibitem[{Zhang et~al.(2019{\natexlab{b}})Zhang, Kishore, Wu, Weinberger, and
  Artzi}]{zhang2019bertscore}
Tianyi Zhang, Varsha Kishore, Felix Wu, Kilian~Q. Weinberger, and Yoav Artzi.
  2019{\natexlab{b}}.
\newblock \href {http://arxiv.org/abs/1904.09675} {Bertscore: Evaluating text
  generation with {BERT}}.
\newblock \emph{CoRR}, abs/1904.09675.

\bibitem[{Zhang et~al.(2018)Zhang, Lapata, Wei, and
  Zhou}]{zhang-etal-2018-neural}
Xingxing Zhang, Mirella Lapata, Furu Wei, and Ming Zhou. 2018.
\newblock \href {https://doi.org/10.18653/v1/D18-1088} {Neural latent
  extractive document summarization}.
\newblock In \emph{Proceedings of the 2018 Conference on Empirical Methods in
  Natural Language Processing}, pages 779--784, Brussels, Belgium. Association
  for Computational Linguistics.

\bibitem[{Zhang et~al.(2019{\natexlab{c}})Zhang, Wei, and
  Zhou}]{zhang-etal-2019-hibert}
Xingxing Zhang, Furu Wei, and Ming Zhou. 2019{\natexlab{c}}.
\newblock \href {https://doi.org/10.18653/v1/P19-1499} {{HIBERT}: Document
  level pre-training of hierarchical bidirectional transformers for document
  summarization}.
\newblock In \emph{Proceedings of the 57th Annual Meeting of the Association
  for Computational Linguistics}, pages 5059--5069, Florence, Italy.
  Association for Computational Linguistics.

\bibitem[{Zhao et~al.(2019)Zhao, Peyrard, Liu, Gao, Meyer, and
  Eger}]{zhao-etal-2019-moverscore}
Wei Zhao, Maxime Peyrard, Fei Liu, Yang Gao, Christian~M. Meyer, and Steffen
  Eger. 2019.
\newblock \href {https://doi.org/10.18653/v1/D19-1053} {{M}over{S}core: Text
  generation evaluating with contextualized embeddings and earth mover
  distance}.
\newblock In \emph{Proceedings of the 2019 Conference on Empirical Methods in
  Natural Language Processing and the 9th International Joint Conference on
  Natural Language Processing (EMNLP-IJCNLP)}, pages 563--578, Hong Kong,
  China. Association for Computational Linguistics.

\bibitem[{Zhong et~al.(2020)Zhong, Liu, Chen, Wang, Qiu, and
  Huang}]{zhong-etal-2020-extractive}
Ming Zhong, Pengfei Liu, Yiran Chen, Danqing Wang, Xipeng Qiu, and Xuanjing
  Huang. 2020.
\newblock \href {https://doi.org/10.18653/v1/2020.acl-main.552} {Extractive
  summarization as text matching}.
\newblock In \emph{Proceedings of the 58th Annual Meeting of the Association
  for Computational Linguistics}, pages 6197--6208, Online. Association for
  Computational Linguistics.

\bibitem[{Zhong et~al.(2019)Zhong, Liu, Wang, Qiu, and
  Huang}]{zhong-etal-2019-searching}
Ming Zhong, Pengfei Liu, Danqing Wang, Xipeng Qiu, and Xuanjing Huang. 2019.
\newblock \href {https://doi.org/10.18653/v1/P19-1100} {Searching for effective
  neural extractive summarization: What works and what{'}s next}.
\newblock In \emph{Proceedings of the 57th Annual Meeting of the Association
  for Computational Linguistics}, pages 1049--1058, Florence, Italy.
  Association for Computational Linguistics.

\bibitem[{Zhou et~al.(2018)Zhou, Yang, Wei, Huang, Zhou, and
  Zhao}]{zhou-etal-2018-neural-document}
Qingyu Zhou, Nan Yang, Furu Wei, Shaohan Huang, Ming Zhou, and Tiejun Zhao.
  2018.
\newblock \href {https://doi.org/10.18653/v1/P18-1061} {Neural document
  summarization by jointly learning to score and select sentences}.
\newblock In \emph{Proceedings of the 56th Annual Meeting of the Association
  for Computational Linguistics (Volume 1: Long Papers)}, pages 654--663,
  Melbourne, Australia. Association for Computational Linguistics.

\end{thebibliography}
\bibliographystyle{acl_natbib}

\appendix
\section{Packages}
For evaluation metrics, we used the following packages:
\begin{itemize}
    \item For ROUGE metrics~\cite{lin-hovy-2003-automatic}, we used the public \emph{rouge-score} package from Google Research: \\ \url{https://github.com/google-research/google-research/tree/master/rouge}
    \item For METEOR~\cite{lavie-agarwal-2007-meteor}, we used the public Natural Language Toolkit:\\
    \url{https://www.nltk.org/_modules/nltk/translate/meteor_score.html}
    \item For BERTScore~\cite{zhang2019bertscore}, we used the public \emph{datasets} package from Huggingface:\\
    \url{https://huggingface.co/metrics/bertscore}
\end{itemize}

\section{Additional Comparison with More Summarization Systems}
\label{sec:additional}
We present the same results in Table~\ref{tab:result} with additional systems in Table~\ref{tab:additional}. Table~\ref{tab:additional} also shows that about half of the listed works employ human evaluation to support the effectiveness of summarization systems.

\begin{landscape}
\begin{table}[t]
\scriptsize
\centering
\begin{tabularx}{\linewidth}{Xllllllllll}
\hline
System          & ROUGE-1    & ROUGE-2   & ROUGE-L   & Average R-Rank & ROUGE-A.M. & ROUGE-G.M. & METEOR   & BERTScore   & Average Rank & Human Eval \\
\hline
Pointer-generator   + coverage ~\citealp{see-etal-2017-get} & 39.53 (34) & 17.28 (33) & 36.38 (33) & 33.33 & 31.06 & 29.18 & 33.1 (16) & 86.44 (15) & 26.20 &     \\
SummaRuNNer                    ~\citealp{nallapati2017summarunner} & 39.6  (33) & 16.2  (34) & 35.3  (34) & 33.67 & 30.37 & 28.29 &          &           & 33.67 &     \\
Pointer + EntailmentGen        ~\citealp{guo-etal-2018-soft} & 39.81 (32) & 17.64 (31) & 36.54 (31) & 31.33 & 31.33 & 29.50 &          &           & 31.33 & yes \\
REFRESH                        ~\citealp{narayan-etal-2018-ranking} & 40.00 (31) & 18.20 (25) & 36.60 (30) & 28.67 & 31.60 & 29.87 & \textbf{43.2} (1) & 87.15 (14) & 20.20 & yes \\
ML+RL ROUGE                    ~\citealp{kryscinski-etal-2018-improving} & 40.19 (30) & 17.38 (32) & 37.52 (25) & 29.00 & 31.70 & 29.70 &          &           & 29.00 & yes \\
                               ~\citealp{li-etal-2018-improving-neural} & 40.30 (29) & 18.02 (27) & 37.36 (26) & 27.33 & 31.89 & 30.05 &          &           & 27.33 & yes \\
ROUGESal+Ent RL                ~\citealp{pasunuru-bansal-2018-multi} & 40.43 (28) & 18.00 (28) & 37.10 (28) & 28.00 & 31.84 & 30.00 &          &           & 28.00 &     \\
RL + pg + cbdec                ~\citealp{jiang-bansal-2018-closed} & 40.66 (27) & 17.87 (30) & 37.06 (29) & 28.67 & 31.86 & 29.97 &          &           & 28.67 & yes \\
end2end w/ inconsistency loss  ~\citealp{hsu-etal-2018-unified} & 40.68 (26) & 17.97 (29) & 37.13 (27) & 27.33 & 31.93 & 30.05 &          &           & 27.33 & yes \\
Latent                         ~\citealp{zhang-etal-2018-neural} & 41.05 (25) & 18.77 (21) & 37.54 (24) & 23.33 & 32.45 & 30.70 &          &           & 23.33 &     \\
Bottom-Up Summarization        ~\citealp{gehrmann-etal-2018-bottom} & 41.22 (24) & 18.68 (24) & 38.34 (19) & 22.33 & 32.75 & 30.91 & 34.2 (15) & 87.71 (11) & 18.60 &     \\
EditNet                        ~\citealp{moroshko-etal-2019-editorial} & 41.42 (23) & 19.03 (19) & 38.36 (18) & 20.00 & 32.94 & 31.15 &          &           & 20.00 &     \\
rnn-ext + RL                   ~\citealp{chen-bansal-2018-fast} & 41.47 (22) & 18.72 (22) & 37.76 (22) & 22.00 & 32.65 & 30.83 & 36.7 (13) & 87.37 (13) & 18.40 & yes \\
BanditSum                      ~\citealp{dong-etal-2018-banditsum} & 41.50 (21) & 18.70 (23) & 37.60 (23) & 22.33 & 32.60 & 30.79 & 39.2 (9) & 87.41 (12) & 17.60 & yes \\
                               ~\citealp{li-etal-2018-improving} & 41.54 (20) & 18.18 (26) & 36.47 (32) & 26.00 & 32.06 & 30.20 &          &           & 26.00 & yes \\
NeuSUM                         ~\citealp{zhou-etal-2018-neural-document} & 41.59 (19) & 19.01 (20) & 37.98 (20) & 19.67 & 32.86 & 31.08 & 39.9 (7) & 88.18 (5) & 14.20 & yes \\
DCA                            ~\citealp{celikyilmaz-etal-2018-deep} & 41.69 (18) & 19.47 (18) & 37.92 (21) & 19.00 & 33.03 & 31.34 &          &           & 19.00 & yes \\
Two-Stage + RL                 ~\citealp{zhang-etal-2019-pretraining} & 41.71 (17) & 19.49 (17) & 38.79 (17) & 17.00 & 33.33 & 31.59 & 35.3 (14) & 87.97 (6) & 14.20 &     \\
HIBERT                         ~\citealp{zhang-etal-2019-hibert} & 42.37 (16) & 19.95 (12) & 38.83 (16) & 14.67 & 33.72 & 32.02 &          &           & 14.67 & yes \\
PNBERT                         ~\citealp{zhong-etal-2019-searching} & 42.69 (15) & 19.60 (16) & 38.85 (15) & 15.33 & 33.71 & 31.91 & 40.3 (6) & 87.73 (9) & 12.20 &     \\
BERT-ext + RL                  ~\citealp{bae-etal-2019-summary} & 42.76 (14) & 19.87 (13) & 39.11 (14) & 13.67 & 33.91 & 32.15 &          &           & 13.67 & yes \\
UniLM                          ~\citealp{dong2019unified} & 43.33 (12) & 20.21 (11) & 40.51 (11) & 11.33 & 34.68 & 32.86 & 38.6 (10) & 88.51 (4) & 9.60  &     \\
T5                             ~\citealp{raffel2019exploring} & 43.52 (11) & \underline{\underline{21.55}} (3) & 40.69 (8) & 7.33  & 35.25 & 33.67 & 38.6 (10) & \underline{88.66} (2) & 6.80  &     \\
DiscoBERT                      ~\citealp{xu-etal-2020-discourse} & 43.77 (10) & 20.85 (8) & 40.67 (9) & 9.00  & 35.10 & 33.36 &          &           & 9.00  & yes \\
BertSum                        ~\citealp{liu-lapata-2019-text} & 43.85 (9) & 20.34 (10) & 39.90 (12) & 10.33 & 34.70 & 32.89 &          &           & 10.33 &     \\
BART                           ~\citealp{lewis-etal-2020-bart} & 44.16 (8) & 21.28 (5) & 40.90 (7) & 6.67  & 35.45 & 33.75 & 40.5 (4) & \underline{\underline{88.62}} (3) & 5.40  & yes \\
PEGASUS                        ~\citealp{zhang2020pegasus} & 44.17 (7) & 21.47 (4) & 41.11 (6) & 5.67  & 35.58 & 33.91 &          &           & 5.67  &     \\
HeterGraph                     ~\citealp{wang-etal-2020-heterogeneous} & 42.95 (13) & 19.76 (15) & 39.23 (13) & 13.67 & 33.98 & 32.17 & 39.7 (8) &           & 12.25 &     \\
ProphetNet                     ~\citealp{qi-etal-2020-prophetnet} & 44.20 (6) & 21.17 (6) & 41.30 (5) & 5.67  & 35.56 & 33.81 &          &           & 5.67  &     \\
MatchSum                       ~\citealp{zhong-etal-2020-extractive} & 44.41 (5) & 20.86 (7) & 40.55 (10) & 7.33  & 35.27 & 33.49 & \underline{41.4} (2) & 87.72 (10) & 6.80  &     \\
Gsum                           ~\citealp{dou-etal-2021-gsum} & 45.94 (4) & \textbf{22.32} (1) & 42.48 (4) & \underline{3.00}  & \underline{\underline{36.91}} & \underline{35.18} &          &           & \underline{3.00}  & yes \\
SimCLS                         ~\citealp{liu-liu-2021-simcls} & \underline{\underline{46.67}} (3) & \underline{22.15} (2) & \underline{\underline{43.54}} (3) & \textbf{2.67}  & \underline{37.45} & \textbf{35.57} & 40.5 (4) & \textbf{88.85} (1) & \textbf{2.60}  &     \\
\hline
Scrambled code + broken                        & \underline{46.71} (2) & 20.39 (9) & \underline{43.56} (2) & \underline{\underline{4.33}}  & 36.89 & 34.62 & 37.5 (12) & 87.8  (7) & 6.40  &     \\
Scrambled code + broken (alter)                     & \textbf{48.18} (1) & 19.84 (14) & \textbf{45.35} (1) & 5.33  & \textbf{37.79} & \underline{\underline{35.13}} & \underline{\underline{40.6}} (3) & 87.8  (7) & \underline{\underline{5.20}}  &    
           \\
\hline
\end{tabularx}
\caption{ROUGE, METEOR, and BERTScore of various summarizers on the CNNDM test set. Ranking of each number in each column is indicated in parentheses. We calculate the average of the ranking, and the smaller the number, the better the ranking. The arithmetic mean (A.M.) and geometric mean (G.M.) of ROUGE-1/2/L obtained by each system (each row) are computed. The \textbf{best score} in each column is in bold, the \underline{runner-up} is underlined, and the \underline{\underline{second runner-up}} is underlined with two lines. Our attack system is compared with well-known summarizers from the past five years. The alternative version (last row) of our system changes $C$ in Algorithm~\ref{alg:b2s} from 3 to 2.}
\label{tab:additional}
\end{table}
\end{landscape}

\end{document}